\documentclass[10pt,twocolumn,letterpaper]{article}

\usepackage{graphicx}
\usepackage{cvpr}              

\usepackage[dvipsnames]{xcolor}

\definecolor{cvprblue}{rgb}{0.21,0.49,0.74}
\definecolor{verylightgray}{rgb}{0.93,0.93,0.93}
\usepackage[pagebackref,breaklinks,colorlinks,citecolor=cvprblue]{hyperref}
\usepackage{colortbl}
\definecolor{babyblue}{rgb}{0.54, 0.81, 0.94}
\definecolor{bisque}{rgb}{1.0, 0.89, 0.77}
\definecolor{bshade}{rgb}{0.55,0.75,0.95}
\usepackage{times}
\usepackage{epsfig}
\usepackage{graphicx}
\usepackage{amsmath}
\usepackage{amssymb}
\usepackage{booktabs}
\usepackage{enumitem}
\usepackage{pifont}
\usepackage{algorithm}
\usepackage{algpseudocode}
\usepackage{multirow}
\usepackage{blindtext}
\usepackage{subcaption}
\usepackage{array}
\usepackage{bbding}
\usepackage{pifont}
\usepackage[numbers]{natbib}
\usepackage[export]{adjustbox}
\usepackage{marginnote}
\usepackage{rotating}
\usepackage{tabularx}
\algnewcommand\algorithmicforeach{\textbf{for each}}
\algdef{S}[FOR]{ForEach}[1]{\algorithmicforeach\ #1\ \algorithmicdo}
\usepackage{color}

\newcommand{\cc}{\color[rgb]{0,0.6,0.3}\checkmark}
\newcommand{\xx}{\color[rgb]{0.6,0,0}{\ding{55}}}
\newcommand{\cmark}{\ding{51}}
\newcommand{\xmark}{\ding{55}}

\usepackage{array}

\newcolumntype{C}[1]{>{\PreserveBackslash\centering}p{#1}}
\newcommand{\PreserveBackslash}[1]{\let\temp=\\#1\let\\=\temp}
\newcolumntype{C}[1]{>{\PreserveBackslash\centering}p{#1}}
\newcolumntype{R}[1]{>{\PreserveBackslash\raggedleft}p{#1}}
\newcolumntype{L}[1]{>{\PreserveBackslash\raggedright}p{#1}}

\newcommand{\fref}[1]{Figure~\ref{#1}}
\newcommand{\sref}[1]{Sec.~\ref{#1}}
\newcommand{\tref}[1]{Table~\ref{#1}}

\definecolor{Gray}{gray}{0.90}
\newcolumntype{g}{>{\columncolor{Gray}}c}
\definecolor{ffe1da}{RGB}{255,225,218}
\definecolor{F7E0D5}{RGB}{247,224,213}
\definecolor{darkF7E0D5}{RGB}{209,154,128}
\colorlet{Light}{White!0!F7E0D5}

\definecolor{Dark}{rgb}{0,0,0}

\definecolor{OutdoorDark}{rgb}{0,.5,0}
\definecolor{comment}{rgb}{0.6, 0.4, 0.8}
\definecolor{IndoorDark}{rgb}{0,0.3,0.8}
\definecolor{SubTDark}{rgb}{0.5,.27,0.11}
\definecolor{AerialDark}{rgb}{.5,.0,.5}
\definecolor{UnderWaterDark}{rgb}{0.16, 0.46, 0.81}
\colorlet{Outdoor}{OutdoorDark!20!white}
\colorlet{Indoor}{IndoorDark!20!white}
\colorlet{SubT}{SubTDark!20!white}
\colorlet{Aerial}{AerialDark!20!white}
\colorlet{UnderWater}{UnderWaterDark!20!white}
\colorlet{OutdoorLight}{OutdoorDark!70!white}
\colorlet{IndoorLight}{IndoorDark!70!white}
\colorlet{SubTLight}{SubTDark!70!white}
\colorlet{AerialLight}{AerialDark!70!white}
\colorlet{UnderWaterLight}{UnderWaterDark!70!white}

\definecolor{tabfirst}{rgb}{1, 0.85, 0.7}

\newcommand{\urban}[1]{\textbf{\textcolor{OutdoorDark}{Urban}}}
\newcommand{\indoor}[1]{\textbf{\textcolor{IndoorDark}{Indoor}}}
\newcommand{\aerial}[1]{\textbf{\textcolor{AerialDark}{Aerial}}}
\newcommand{\subt}[1]{\textbf{\textcolor{SubTDark}{SubT}}}
\newcommand{\degraded}[1]{\textbf{\textcolor{SubTDark}{Degraded}}}
\newcommand{\underwater}[1]{\textbf{\textcolor{UnderWaterDark}{Underwater}}}

\def\paperID{16421} 
\def\confName{CVPR}
\def\confYear{2024}

\title{SubT-MRS Dataset: Pushing SLAM Towards All-weather Environments \\
\normalfont \color{brown}{\href{https://superodometry.com/datasets}{\textbf{https://superodometry.com/datasets}}}
}

\author{\fontsize{11pt}{11pt}\selectfont Shibo Zhao\thanks{Corresponding author: shiboz@andrew.cmu.edu}  \textsuperscript{1}, Yuanjun Gao\textsuperscript{1}, Tianhao Wu\textsuperscript{1}, Damanpreet Singh\textsuperscript{1}, Rushan Jiang\textsuperscript{1}, Haoxiang Sun\textsuperscript{1},\\\fontsize{11pt}{11pt}\selectfont Mansi Sarawata\textsuperscript{1}, Yuheng Qiu\textsuperscript{1}, Warren Whittaker\textsuperscript{1}, Ian Higgins\textsuperscript{1}, Yi Du\textsuperscript{2}, Shaoshu Su\textsuperscript{2}, Can Xu\textsuperscript{1}, John Keller\textsuperscript{1},\\\fontsize{11pt}{11pt}\selectfont Jay Karhade\textsuperscript{1}, Lucas Nogueira\textsuperscript{1}, Sourojit Saha\textsuperscript{1}, Ji Zhang\textsuperscript{1}, Wenshan Wang\textsuperscript{1}, Chen Wang\textsuperscript{2}, Sebastian Scherer\textsuperscript{1} \vspace{1.5 mm}\\
{\fontsize{11pt}{11pt}\selectfont \textsuperscript{1}Carnegie Mellon University \quad } 
{\fontsize{11pt}{11pt}\selectfont \textsuperscript{2}University at Buffalo}
}

\begin{document}

\maketitle

\begin{figure*}[!ht]
    \centering
    \includegraphics[width=1.0\linewidth]{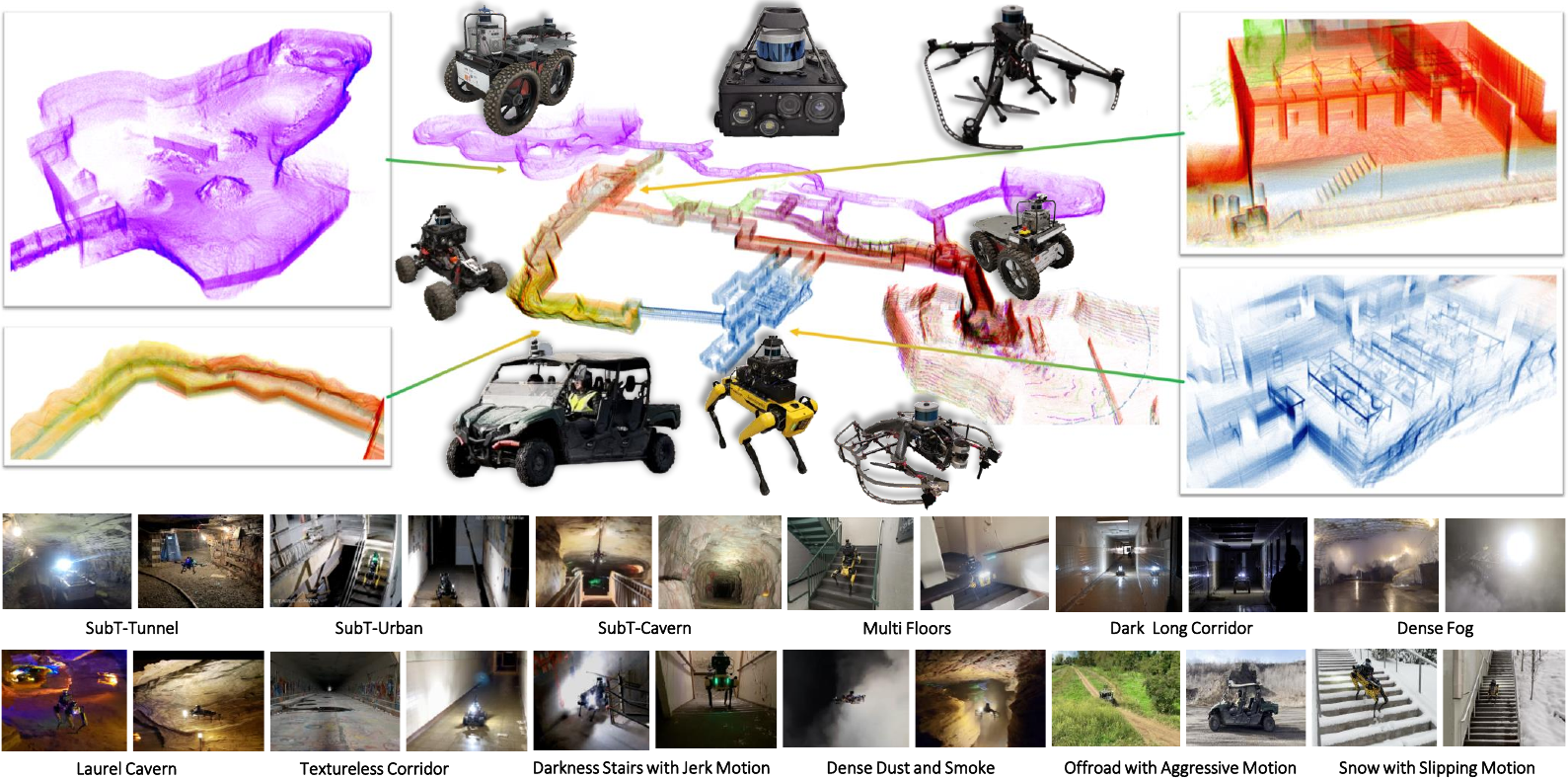}
    \caption{Dense reconstruction from the SubT-MRS dataset, achieved through collaboration with diverse robots equipped with multimodal sensors. Colors represent different challenging environments (tunnels, caves, urban, confined spaces) captured by various robot types (aerial, legged, wheeled). The bottom section showcases a gallery with diverse visual, LiDAR, and mixed degradations.}
    \label{fig:ground_truth}
\end{figure*}

\vspace{-1cm}

\begin{abstract}

Simultaneous localization and mapping (SLAM) is a fundamental task for numerous applications such as autonomous navigation and exploration. Despite many SLAM datasets have been released, current SLAM solutions still struggle to have sustained and resilient performance. One major issue is the absence of high-quality datasets including diverse all-weather conditions and a reliable metric for assessing robustness. This limitation significantly restricts the scalability and generalizability of SLAM technologies, impacting their development, validation, and deployment. To address this problem, we present SubT-MRS, an extremely challenging real-world dataset designed to push SLAM towards all-weather environments to pursue the most robust SLAM performance.  It contains multi-degraded environments including over 30 diverse scenes such as structureless corridors, varying lighting conditions, and perceptual obscurants like smoke and dust; multimodal sensors such as LiDAR, fisheye camera, IMU, and thermal camera; and multiple locomotions like aerial, legged, and wheeled robots. We developed accuracy and robustness evaluation tracks for SLAM and introduced novel robustness metrics. Comprehensive studies are performed, revealing new observations, challenges, and opportunities for future research. 

\end{abstract}


\section{Introduction}
\label{sec:intro}

Simultaneous Localization and Mapping (SLAM) is essential in robotics since it provides foundational perception and spatial awareness, and enables machines to understand and interact with the physical world in real-time. 
Therefore, it has a wide range of applications such as autonomous driving and space exploration.
Despite significant advancements in both geometric \cite{vins-mono, liosam2020shan, orb-slam3} and data-driven SLAM methods \cite{teed2021droid,wang2021tartanvo}, existing solutions remain fragile in challenging conditions.
One main reason is that existing SLAM algorithms are often developed and evaluated with datasets from controlled environments \cite{Burri25012016, Delmerico19icra, Geiger2012CVPR, helmberger2022hilti, ramezani2020newer, pfrommer2017penncosyvio, schubert2018tum, zuniga2020vi}.
For example, KITTI \cite{Geiger2012CVPR, helmberger2022hilti} dataset is mostly collected in sunny weather and EuROC-MAV dataset \cite{Burri25012016} is collected in small well-lit rooms.
These datasets, unfortunately, fail to capture the challenges in real-world scenarios, hindering the development of robust SLAM solutions.

To bridge this gap and push SLAM towards all-weather environments, we present an extremely challenging dataset, SubT-MRS, including scenarios featuring various sensor degradation, aggressive locomotions, and extreme-weather conditions.
The SubT-MRS dataset comprises 3 years of data from the DARPA Subterranean (SubT) Challenge \cite{darpasubt} (2019$-$2021) and extends with an additional 2 years of diverse environments (2022$-$2023), containing mixed indoor and outdoor settings, including long corridors, off-road scenario, tunnels, caves, deserts, forests, and bushlands. 
Cumulatively, this forms a 5-year dataset encompassing over 500 hours and 100 kilometers of terrain subjected to \textbf{multimodal sensors} including LiDAR, fisheye cameras, depth cameras, thermal cameras, and IMU; \textbf{heterogeneous platforms} including RC cars, legged robots, aerial robots, and wheeled robots; and \textbf{extreme obscurant conditions} such as dense fog, dust, smoke, and heavy snow.

Additionally, we find there is no well-established metric to evaluate the robustness of a SLAM system. Existing evaluation metrics such as absolute trajectory error (ATE) \cite{Burri25012016} are not representative of the actual performance in robotic applications.
We argue that to ensure the safety control of a robot, SLAM system evaluation should not only focus on poses but also its velocity.
Taking robot localization as an example, while a momentary pose error spike may only slightly undermine the overall ATE, however, it can lead to a catastrophic crash in an aerial robotic platform.
To effectively gauge the performance of a SLAM algorithm, we introduce a new \textit{robustness} metric to evaluate the reliability of the SLAM system, particularly examining smoothness and accuracy of velocity estimation.

Lastly, we perform extensive experiments using proposed degradation datasets to benchmark visual and LiDAR SLAM algorithms. 
These experiments identify the limitations of the existing SLAM systems and evaluate their robustness across the degradation. 
Our contributions include:
\begin{itemize}[noitemsep,topsep=0pt]
     \item \textbf{All-weather Environments}~~To push SLAM toward all-weather environments, we introduce SubT-MRS, an extremely challenging dataset. Spanning five years, it comprises over 500 hours and 100 kilometers of accurately measured trajectories.
     To the best of our knowledge, SubT-MRS is the first real-world dataset that specifically addresses failure scenarios of SLAM by incorporating a variety of degraded conditions, multiple robotic platforms, and diverse sets of multimodal sensors. 
    \item \textbf{Robustness Metric}~~To evaluate the actual performance running on robots, we propose a \textit{robustness} metric, which, to the best of our knowledge, is the first metric evaluating the reliability, safety, and resilience of SLAM.
    \item \textbf{SLAM Challenge}~~We provide a comprehensive benchmark\footnote{SubT-MRS dataset is completely real-world data. Simulated data in the benchmark is to ensure evaluation completeness} based on the SubT-MRS dataset by conducting ICCV'23 SLAM Challenge. Our evaluation reveals the limitations of state-of-the-art visual and LiDAR SLAM solutions. Furthermore, we conduct extensive experiments to verify robustness metric against various sensor degradation and length of trajectories. 
\end{itemize}

\begin{table*}[h!]
	\small
	\begin{center}
    \resizebox{\textwidth}{!}{
 
		\begin{tabular}{l|cccc|ccccc|cccc}
			\toprule
			\multicolumn{1}{c|}{\multirow{3}{*}{\textbf{Dataset}}}  & \multicolumn{4}{c|}{\textbf{Multi-Spectrum}}   & \multicolumn{5}{c|}{\textbf{Multi-Degradation}}  & \multicolumn{4}{c}{\textbf{Multi-Robot}} \\ 
			\multicolumn{1}{c|}{} & \multicolumn{1}{l}{\multirow{2}{*}{Camera}} & \multicolumn{1}{l}{\multirow{2}{*}{IMU}} & \multicolumn{1}{l}{\multirow{2}{*}{LiDAR/Depth}} & \multicolumn{1}{l|}{\multirow{2}{*}{Thermal}} & \multicolumn{1}{l}{\multirow{2}{*}{Illumination}} & \multicolumn{1}{l}{\multirow{1}{*}{Snow}} & \multicolumn{1}{l}{\multirow{2}{*}{Structureless}} & \multicolumn{1}{l}{\multirow{2}{*}{SubT}}     & \multicolumn{1}{l|}{\multirow{2}{*}{Aggressive Motion}} & \multicolumn{1}{l}{\multirow{2}{*}{Vehicle}} & \multicolumn{1}{l}{\multirow{2}{*}{Drone}} & \multicolumn{1}{l}{\multirow{2}{*}{Legged}} & \multicolumn{1}{l}{\multirow{2}{*}{Handheld}} \\
            \multicolumn{1}{c|}{} & \multicolumn{1}{c}{} & \multicolumn{1}{c}{} & \multicolumn{1}{c}{} & \multicolumn{1}{c|}{} & \multicolumn{1}{c}{} & \multicolumn{1}{c}{Smoke} & \multicolumn{1}{c}{} & \multicolumn{1}{c}{} & \multicolumn{1}{c|}{} & \multicolumn{1}{c}{} & \multicolumn{1}{c}{} & \multicolumn{1}{c}{}& \multicolumn{1}{c}{}  \\ \midrule
			EuRoC MAV \cite{Burri25012016}          & \cc & \cc & \xx & \xx & \cc & \xx & \xx & \xx & \xx & \xx & \cc & \xx & \xx \\
			PennCOSYVIO~\cite{pfrommer2017penncosyvio} & \cc & \cc & \xx & \xx & \xx & \xx & \xx & \xx & \xx & \xx & \xx & \xx & \cc \\
			TUM VIO~\cite{schubert2018tum}           & \cc & \cc & \xx & \xx & \cc  & \xx & \xx & \xx & \xx & \xx & \xx & \xx & \cc\\
			UZH-FPV~\cite{Delmerico19icra}          & \cc & \cc & \xx & \xx & \xx & \xx & \xx & \xx & \cc & \xx & \cc & \xx & \xx\\
		
           
			College Dataset~\cite{zhang2021multi}   & \cc & \cc & \cc & \xx & \xx & \xx & \xx & \xx & \xx & \xx & \xx & \xx & \cc\\
			RobotCar \cite{maddern20171}            & \cc & \cc & \cc & \xx & \cc & \cc & \xx & \xx & \xx & \cc & \xx & \xx & \xx\\
            
            UMA VI~\cite{zuniga2020vi}              & \cc & \cc & \xx & \xx & \cc & \xx & \cc & \xx & \xx & \xx & \xx & \xx & \cc \\
            UMich~\cite{carlevaris2016university}   & \cc & \cc & \cc & \xx & \cc & \xx & \xx & \xx & \xx & \cc & \xx & \xx & \xx \\ 
            KITTI \cite{geiger2013vision}   		& \cc & \cc & \cc & \xx & \xx & \xx & \cc & \xx & \cc & \cc & \xx & \xx & \xx  \\
		Virtual KITTI \cite{gaidon2016virtual}  & \cc & \cc & \cc & \xx & \cc & \xx & \cc & \xx &\cc   & \cc & \xx & \xx & \xx\\
            {DARPA SubT} ~ \cite{rogers2021darpa}                                    & \cc & \cc & \cc & \xx & \cc & \xx & \cc & \cc & \xx & \cc  & \xx & \xx &\xx  \\
              
            Hilti SLAM~\cite{helmberger2022hilti}   & \cc & \cc & \cc & \xx & \cc & \xx & \cc & \xx & \xx & \xx & \xx & \xx & \cc \\

            {M3ED} ~\cite{chaney2023m3ed}  & \cc & \cc & \cc & \xx & \cc & \xx & \xx & \xx & \xx & \cc    &  \cc  & \cc   & \xx   \\
            
            TartanAir~\cite{tartanair2020iros}      & \cc & \cc & \cc & \xx & \cc & \cc & \cc & \xx & \cc &  \xx  &  \cc  &  \xx  &  \xx  \\
            
			\textbf{SubT-MRS (Ours)}                                    & \cc & \cc & \cc & \cc & \cc & \cc & \cc & \cc & \cc & \cc  & \cc & \cc & \cc \\
        \bottomrule
		\end{tabular} 
        }
	\vspace{-0.2in}
	\end{center}
	\caption{Comparison of SLAM datasets on multi-sensors, multi robots, and multi degradation.}

	\label{Tab:dataset_compare}
\end{table*}
\section{Related Work} 
\label{sec:related_work}

\paragraph{Multimodal Sensors}
Multimodal sensor datasets are crucial for the development of robust SLAM systems, as single sensor modalities are not comprehensive for all scenarios. Existing datasets typically focus on a limited range of modalities, such as monocular or stereo cameras \cite{Burri25012016,wenzel2020fourseasons}, LiDAR \cite{geiger2013vision}, event cameras \cite{mueggler2017event}, and depth cameras \cite{sturm12iros}. The KITTI dataset \cite{geiger2013vision}, although covering most modalities, is geared towards on-road scenarios and lacks thermal cameras, limiting its applicability to simple, controlled environments. Conversely, the ViViD++ dataset \cite{lee2022vivid++} incorporates thermal images in outdoor settings but is deficient in hardware synchronization, posing challenges for SLAM system development. The Weichen dataset \cite{dai2021multi} offers thermal images in indoor environments with accurate ground truth poses but is confined to motion capture room settings. In contrast, our SubT-MRS dataset delivers a comprehensive suite of time-synchronized multimodal sensors, including LiDAR, monocular cameras, thermal cameras, depth cameras, and IMU, along with centimeter-level ground truth, catering to a diverse range of research needs.

\paragraph{Heterogeneous Platforms} 
Developing a robust SLAM system necessitates versatility in handling different motion patterns and various robotic platforms.
However, most existing datasets are tailored for single-robot scenarios. While there are a few multi-robot datasets, they typically involve homogeneous platforms \cite{utias,chen2022milestones,feng2022s3e,tian2023resilient}, posing challenges in evaluating SLAM performance across varied robotic platforms.
The AirMuseum dataset \cite{air_museum} includes drones and ground robots but lacks data from legged robots. 
As a comparison, SubT-MRS dataset features diverse robotic platforms including legged robots, aerial robots, and wheeled robots, operating in varied environments and sensor setups.

\paragraph{Extreme Environmental Conditions}
Developing and testing SLAM systems in extreme environmental conditions is crucial for mitigating potential real-world failures. However, most existing SLAM datasets have been primarily limited to single, controlled environments. The TUM-VI \cite{schubert2018tum} and UMA-VI \cite{zuniga2020vi} datasets, being indoor-outdoor visual-inertial datasets, pose challenges due to varying illumination and low-texture environments. The EuRoC MAV \cite{Burri25012016} and UZH-FPV drone racing \cite{Delmerico19icra} datasets, popular in the SLAM community, offer data on aggressive drone movements but usually in consistently lit conditions. The KITTI dataset \cite{geiger2013vision}, a staple in autonomous driving research for its outdoor LiDAR-Visual-Inertial data. The Hilti SLAM dataset \cite{helmberger2022hilti} includes both indoor and outdoor LiDAR-visual-inertial datasets with dynamic lighting and confined spaces. However, both of them focus more on accuracy than on a variety of environmental degradations. TartanAir \cite{tartanair2020iros} covers most of the challenging environments but it is a simulation dataset that will pose the sim-to-real gap.

In contrast, the SubT-MRS dataset provides 30 varied scenes, encompassing a wide array of environments. These include SubT tunnels, urban areas, caverns, multi-floor structures, dark corridors, and foggy conditions. It also features textureless surfaces, dusty and smoky environments, off-road areas with aggressive motion, and snowy terrains prone to slippage. We include challenging lighting conditions such as extreme darkness and overexposure, along with geometrically challenging scenarios like featureless corridors, staircases, and self-similar cave layouts. Additionally, the dataset incorporates perceptual challenges posed by smoke, fog, dust, and various weather conditions.

\section{SubT-MRS Dataset} \label{sec:dataset}

We next present the SubT-MRS dataset from three aspects including its collecting environments and settings, ground truth collection, and the new robustness metric.
A detailed comparison with other datasets is listed in \tref{Tab:dataset_compare}.

\begin{figure*}[!ht]
    \centering
    \includegraphics[width=0.9\linewidth]{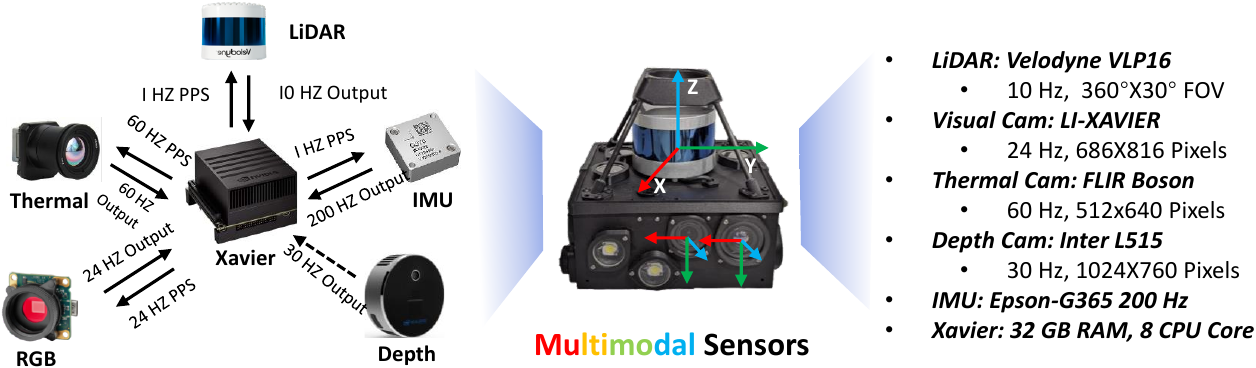}
    \caption{An overview of the sensor pack used in SubT-MRS dataset. It is equipped with a Xavier processing unit with hardware time synchronization for multimodal sensors including LiDAR, fisheye cameras, thermal cameras, depth cameras (option), and an IMU.}
    \label{fig:sensor}
\end{figure*}

\vspace{-0.2 cm}

\subsection{All-weather Environments} 

As mentioned in \sref{sec:related_work}, the SubT-MRS dataset distinguishes itself through its multimodal sensor setups, heterogeneous robotic platforms, and extreme environmental conditions.
This section emphasizes this comprehensive nature pushing SLAM towards all-weather environments.

\subsubsection{Multimodal Sensors}
Multimodal sensors provide a wealth of information for SLAM systems operating in all-weather environments. 
This diversity improves scene understanding and strengthens the system's perception ability. Therefore,
We incorporate diverse sensors to ensure system robustness. 

\paragraph{Sensor Pack}
We embedded 4 Leopard Imaging RGB fisheye cameras, 1 Velodyne puck, 1 Epson M-G365 IMU, 1 FLIR Boson thermal camera, and NVIDIA Jetson AGX Xavier as a sensor pack shown in Figure \ref{fig:sensor}.

\paragraph{Time Synchronization} 
To ensure the overall consistency of the fused data, we meticulously synchronize the sensors' time using the `pulse per second (PPS)' technique.
As illustrated in \fref{fig:sensor}, the IMU, LiDAR, and thermal camera are directly synchronized with the CPU clock, while the four RGB camera are synchronized using an FPGA board. 
Consequently, we can effectively manage the time synchronization gap between each pair of sensors not to exceed 3ms.

\paragraph{Fisheye and Thermal Camera Calibration} Camera calibration plays an essential role in the efficiency of a SLAM system. For calibrating the fisheye cameras, we employed the open-source toolkit Kalibr \cite{rehder2016extending}, focusing on the intrinsic and extrinsic parameters. Specifically, we use the radial-tangential distortion model to rectify the omnidirectional fisheye camera model.
Calibrating thermal cameras, however, poses unique challenges, especially in gathering high-quality thermal data. To tackle this, we set up a 7$\times$9 chessboard, heated by direct sunlight, to generate high-contrast thermal data, ensuring accuracy in the calibration process.

\paragraph{IMU Calibration and Extrinsic Calibration}
For Lidar-IMU extrinsic calibration, we utilize the CAD model to obtain the calibration parameters. 
In the case of Camera-IMU extrinsic calibration, we employ the Kalibr toolbox \cite{rehder2016extending} to estimate the extrinsic matrix.
To estimate the sensors' bias and the random walk noise of the gyroscope and accelerometer, we collected static data from the IMU and calibrated it using an Allan variance-based IMU tool \cite{4404126, imu_tool}.

\begin{figure*}[t]
\centering\includegraphics[width=\textwidth,height=0.9\textheight,keepaspectratio]{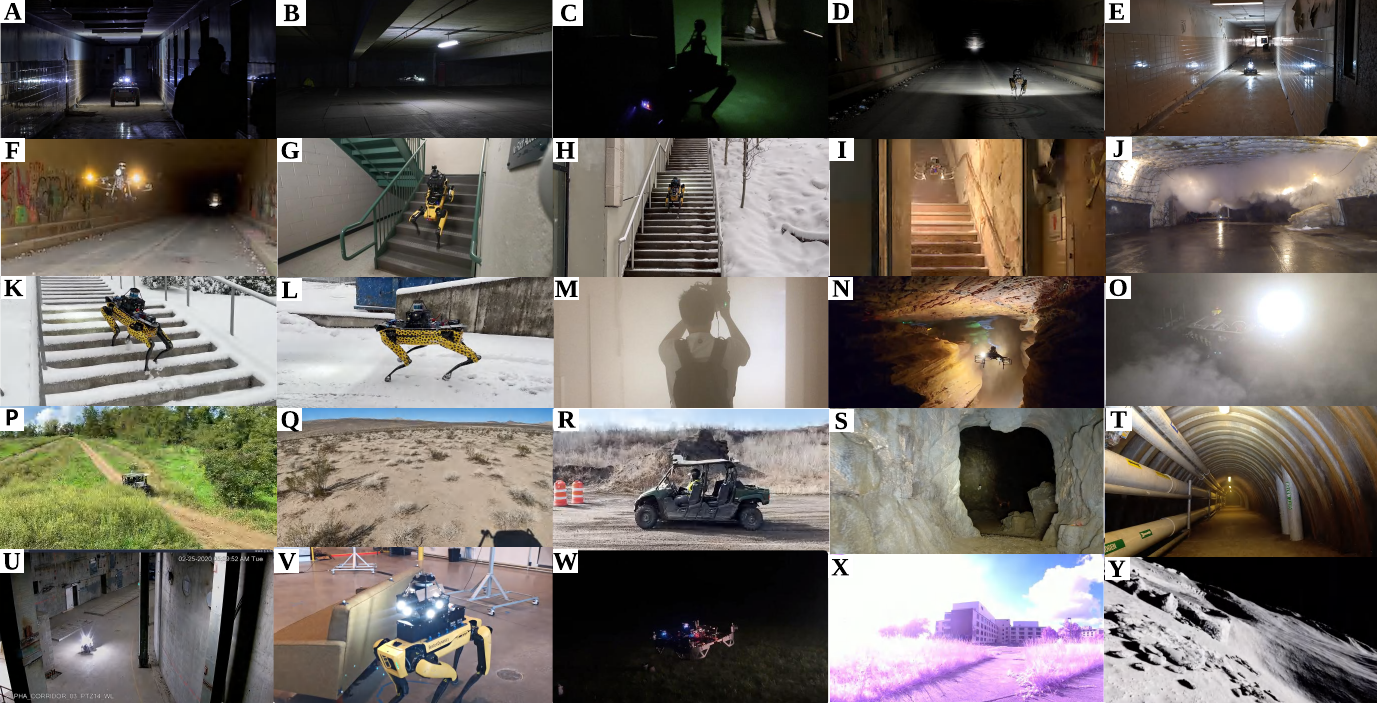}
    \caption{The SubT-MRS datasets were collected across diverse seasons, capturing environments with perceptual challenges such as poor illumination, darkness, and water puddles, where visual sensors falter. They also include geometrically complex areas like long featureless corridors and steep multi-floor structures, challenging LiDAR systems with potential drift. Moreover, these datasets cover conditions with airborne obscurants like dust, fog, snow, and smoke in tough environments, including caves, deserts, long tunnels, and off-road areas.} 
    \label{fig:various_terrains}
\end{figure*}

\begin{figure*}[t]
    \centering
    \includegraphics[width=0.9\textwidth]{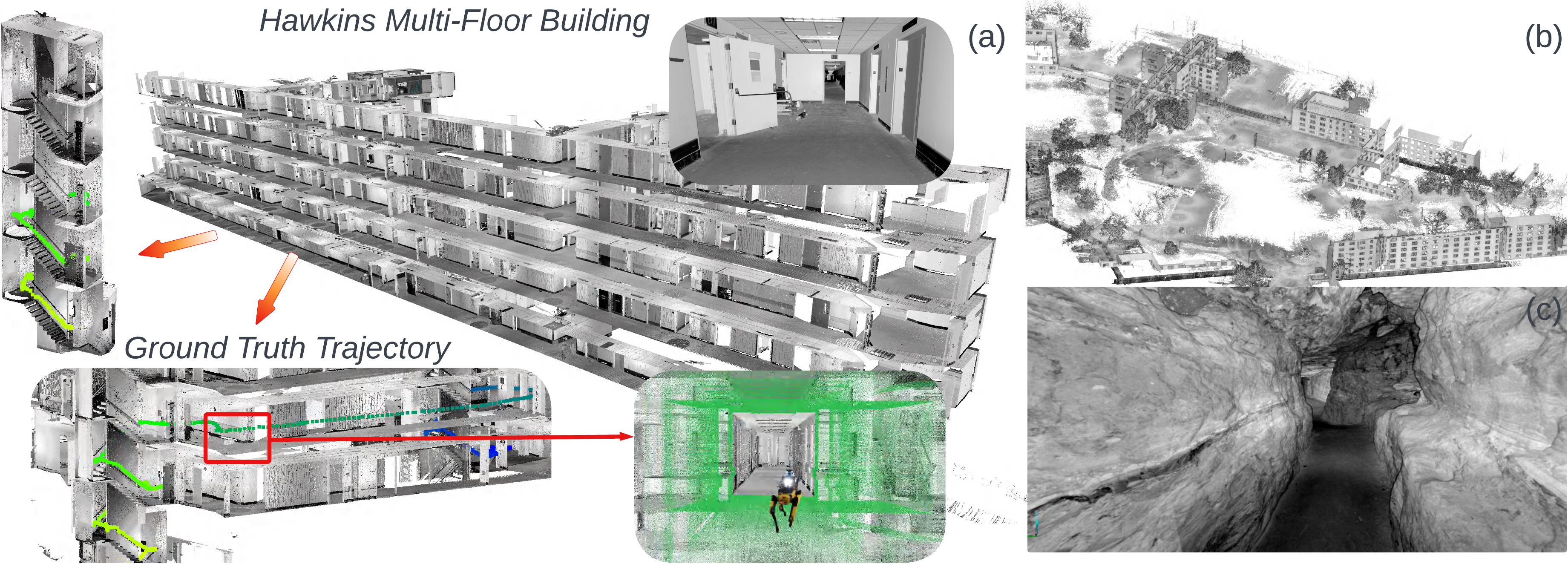}
    \caption{The SubT-MRS dataset facilitates the generation of high-precision ground truth maps and trajectories. Figure (a) shows the ground truth trajectory in multi-floor settings. Figure (b) displays ground truth maps for indoor and outdoor areas, encompassing long corridors, multi-floor structures, and open spaces. Figure (c) features photo-realistic scans based on our ground truth maps in cave environments.} 
    \label{fig:groundtruth_map}
\end{figure*}

\subsubsection{Multi-degraded Environments}
\label{sec:multi-degradation}

SubT-MRS includes multiple challenging environments including visually degraded environments, geometrically degraded environments, and their combination.

\paragraph{Visual Degradation} Poor-quality visual features can significantly hinder the performance of feature extraction processes and disrupt the accuracy of feature matching. Such issues arise in low-light conditions with inconsistent brightness or image noise introduced by air obscurants. SubT-MRS encompasses these challenges and provides a wide range of visual degradation. This includes environments with limited lighting, such as hospital interiors and caves (Figure \ref{fig:various_terrains} A-F), as well as smoky or dusty conditions that cause visual obstruction (Figure \ref{fig:various_terrains} M-N), and snowy areas with reduced visibility (Figure \ref{fig:various_terrains} H, K, and L).

\paragraph{Geometric Degradation} Lack of geometrical features poses significant challenges to LiDAR odometry. The root of this issue often lies in the limited sensing capabilities of LiDAR sensors and the constraints due to their mechanical installation. SubT-MRS captures a variety of environments that exemplify such challenges. It includes long, featureless corridors (\fref{fig:various_terrains} E) and staircases (\fref{fig:various_terrains} C and G). These scenarios illustrate various forms of geometric degradation, contributing to the improvement of LiDAR odometry systems in challenging conditions.

\paragraph{Mixed Degradation} A mixed visual and geometric degradation can further hinder the performance of SLAM systems. Examples in SubT-MRS include long, dimly lit corridors (\fref{fig:various_terrains} A, E, and F), poorly illuminated staircases (\fref{fig:various_terrains} C), and environments affected by snowy weather (\fref{fig:various_terrains} H, K, and L). In these settings, both LiDAR odometry and visual odometry are prone to failure due to the compounded effects of mixed degradation. The inclusion of such scenarios in SubT-MRS is crucial for evaluating and improving the resilience of multimodal SLAM algorithms.

\subsubsection{Heterogeneous Robot Platforms}
Most of current datasets focus on single-robot systems, limiting multi-robot SLAM development as shown in Table \ref{Tab:dataset_compare}. 
To address this, we employed diverse robot platforms, including aerial, legged, and wheeled robots, navigating through various environments from urban campuses to medical facilities and natural terrains like caves. This diversity offers a range of scenarios to enhance SLAM algorithms for effective operation in challenging conditions.

\paragraph{Extrinsic Calibration for Multiple Robots}\label{sec:extrinsic_calib_robots}
To ensure the multi-robot system shares a common coordinate system, we perform the extrinsic calibration process \cite{scherer2022resilient}.
It has 2 steps: First, each robot runs Super Odometry \cite{zhao2021super} to generate its local map using LiDAR data and share it with other robots through a wireless network; Second, the remaining robots identify overlapping regions between their local maps and estimate the extrinsic parameters using GICP \cite{segal2009generalized}.

\setlength{\tabcolsep}{3pt}

\begin{table*}[h]
\setlength{\tabcolsep}{1pt}
\renewcommand{\arraystretch}{1.0}
\centering
\caption{SLAM Challenge Results (Blue shadings indicate rankings of ATE and Robustness Metric; L: LiDAR I: IMU C: Camera)}
\scalebox{0.7}{
\begin{tabular}{c c  c C{2cm} c  c c c c c c c c} 
\toprule
\multirow{2}{*}{\textbf{\#}} &\multirow{2}{*}{\textbf{Team}} & \multirow{2}{*}{\textbf{Method}} & \textbf{ Odometry }& \multirow{2}{*}{\textbf{ Device}} & \multirow{2}{*}{\textbf{ RealTime (s) }}  & \multirow{2}{*}{\textbf{CPU/GPU (\%) }} & \multirow{2}{*}{\textbf{ RAM (GB) }}  & \multirow{2}{*}{\textbf{ ATE$\downarrow$ }} & \multirow{2}{*}{\textbf{ $R_v\uparrow$/$R_w\uparrow$}} & \multicolumn{3}{c}{\textbf{Sensors}}  \\
&  & & \textbf{Type} &  & & & & & & \textbf{ L} & \textbf{I} & \textbf{C}  \\
\midrule
1 & Liu et al   & FAST-LIO2 \cite{xu2022fast}, HBA \cite{liu2023efficient} &  Filter &  Intel i7-9700K  &  51.310 & 98.667 / 0& 4.052 & \colorbox{bshade!110}{\makebox[3em][c]{\textbf{0.588}}}& \colorbox{bshade!110}{\makebox[5em][c]{\textbf{0.517}/\textbf{0.770}}} & $\checkmark$ & $\checkmark$ &  \\ 

2 & Yibin et al    & LIO-EKF \cite{vizzo2023kiss} & Filter  & Intel i7-10700  & \textbf{0.006} & 52.167 / 0  & 0.072 & \colorbox{bshade!80}{\makebox[3em][c]{4.313}} & \colorbox{bshade!40}{\makebox[5em][c]{0.441/0.574}} &  $\checkmark$& $\checkmark$ & \\ 

3 & Weitong et al & FAST-LIO\cite{faster_lio}, Pose Graph\cite{dellaert2017factor} & Filter  & Intel Xeon(R)E3-1240v5 & 0.125 & 22.63 / 0 & 4.305 & \colorbox{bshade!60}{\makebox[3em][c]{0.663}} & \colorbox{bshade!80}{\makebox[5em][c]{0.473/0.747}} & $\checkmark$  & $\checkmark$ & \\

4 & Kim et al & FAST-LIO2\cite{fastlio}, Point-LIO\cite{he2023point}, Quatro\cite{lim2022quatro} & Filter  & Intel i5-12500 & 0.268 & 101.108 / 0  & 55.64 & \colorbox{bshade!40}{\makebox[3em][c]{3.825}} & \colorbox{bshade!60}{\makebox[5em][c]{0.479/0.615}} & $\checkmark$  &$\checkmark$ & \\ 

5 & Zhong et al & DLO\cite{chen2022direct}, Scan-Context++\cite{kim2018scan} & SW Opt  & AMD Ryzen 9 5900x & 0.027 & \textbf{13.289} / 0 & 1.174 & \colorbox{bshade!30}{\makebox[3em][c]{1.209}} & \colorbox{bshade!30}{\makebox[5em][c]{0.276/0.486}} & $\checkmark$  & $\checkmark$& \\
\midrule
1 & Peng  et al & DVI-SLAM \cite{peng2023dvislam} & Learning & Intel i9-12900 & 183.233 & - / 149 & 11 (4) & \colorbox{bshade!110}{\makebox[3em][c]{\textbf{0.547}}} & \colorbox{bshade!110}{\makebox[5em][c]{\textbf{0.473}/\textbf{0.788}}} & & $\checkmark$ & $\checkmark$  \\

2 & Jiang et al & LET-NET\cite{let-net}, VINS-Mono\cite{vins-mono}& Hybrid  & Intel i5-9400 & 0.064 & \textbf{40.35} / 0 & 4.337 & \colorbox{bshade!80}{\makebox[3em][c]{1.093}} & \colorbox{bshade!40}{\makebox[5em][c]{0.078/0.322}} & & $\checkmark$  & $\checkmark$ \\

3 & Thien et al & VR-SLAM\cite{nguyen2023vrslam} &  SW Opt & Intel i9-12900 & 0.142 & 176.44 / 0 & 9.111 & \colorbox{bshade!40}{\makebox[3em][c]{3.037}} & \colorbox{bshade!60}{\makebox[5em][c]{0.083/0.372}} & &   $\checkmark$ &  $\checkmark$\\

4 & Li et al   & ORB-SLAM3\cite{campos2021orb}& SW Opt.  & Intel i7-10700 & \textbf{0.019} & 65.028 / 0 & 0.386 & \colorbox{bshade!30}{\makebox[3em][c]{8.975}} & \colorbox{bshade!80}{\makebox[5em][c]{0.163/0.474}} & & $\checkmark$ &  $\checkmark$\\
\bottomrule
\end{tabular}
}
\label{tab:iccv23}
\end{table*}

\begin{table*}[t]
\centering
\caption{Accuracy Performance on Geometric Degradation. Red numbers represent ATE ranking. * denotes incomplete submissions.}
\scalebox{0.76}{
\begin{tabular}{c|cccccc|ccc|ccc|c}
\toprule

\multirow{2}{*}{\textbf{Team}} & \multicolumn{6}{c|}{\color{Dark} \textbf{Geometric Degradation (Real World)}} &\multicolumn{3}{c|}{\color{Dark} \textbf{Simulation}} &\multicolumn{3}{c|}{\color{Dark} \textbf{Mix Degradation}} & \multirow{2}{*}{\textbf{Average}} \\

\cmidrule(lr{0.75em}){2-7}  \cmidrule(lr{0.75em}){8-10} \cmidrule(lr{0.75em}){11-13} 
 
 & \color{Dark} Urban  & \color{Dark} Tunnel & \color{Dark} Cave & \color{Dark} Nuclear\_1  & \color{Dark} Nuclear\_2  & \color{Dark} {Laurel\_Caverns}   &  \color{Dark}Factory &  \color{Dark}Ocean &   \color{Dark}Sewerage & \color{Dark}Long Corridor & \color{Dark}Multi Floor & \color{Dark}Block Lidar & \\ 

\cmidrule{1-2}  \cmidrule(lr{0.75em}){3-3} \cmidrule(lr{0.75em}){4-4} \cmidrule(lr{0.75em}){5-5} \cmidrule(lr{0.75em}){6-6} \cmidrule(lr{0.75em}){7-7} \cmidrule(lr{0.75em}){8-8} \cmidrule(lr{0.75em}){9-9} \cmidrule(lr{0.75em}){10-10} \cmidrule(l{0.75em}){11-11} \cmidrule(lr{0.75em}){12-12} \cmidrule(lr{0.75em}){13-13} \cmidrule(lr{0.75em}){14-14} 

Liu et al\textsuperscript{\color{Red}\textbf{1}}  & 0.307  & 0.095 & 0.629 & 0.122 & 0.235              & \color{UnderWaterDark}{\textbf{0.260}}      &    \color{UnderWaterDark}{\textbf{0.889}} & 0.757 & \color{UnderWaterDark}{\textbf{0.978}}  & 1.454 & \color{UnderWaterDark}{\textbf{0.401}} & \color{UnderWaterDark}{\textbf{0.934}}  & \color{UnderWaterDark}{\textbf{0.588}}\\

Weitong et al\textsuperscript{\color{Red}\textbf{2}}   & \color{UnderWaterDark}{\textbf{0.26}} & 0.096         & \color{UnderWaterDark}{\textbf{0.617}}           & \color{UnderWaterDark}{\textbf{0.120}}        &  \color{UnderWaterDark}{\textbf{0.222}}       & 0.402        & 0.998  & 0.770    & 1.586  & \color{UnderWaterDark}{\textbf{1.254}} & 0.577 & 1.056 & 0.663 \\

Kim et al\textsuperscript{\color{Red}\textbf{3}}  & 0.331          & \color{UnderWaterDark}{\textbf{0.092}}          & 0.787      & 0.123        & 0.270          & 0.279  & 10.628          & 22.425        & 7.147          & 2.100   & 0.650 & 1.068 & 3.825  \\

Yibin et al\textsuperscript{\color{Red}\textbf{4}} & 1.060          &0.220         & 0.750           & 0.470         & 0.620          & 9.140          & 4.920         & \color{UnderWaterDark}{\textbf{0.280}}         & 24.460  & 2.990 & 5.500 & 1.340 & 4.312  \\



Zhong et al\textsuperscript{\color{Red}\textbf{5}}  & 1.205 & 0.695 & - & 1.175 & 1.72 & 2.08 & \color{UnderWaterDark}{\textbf{0.889}} & 0.778 & 1.13 & -  & - & - & 1.209\textsuperscript{\textbf{*}}  \\
\bottomrule

\multicolumn{1}{c|}{\textbf{Average}}  & 0.633 & 0.240 & 0.696 & 0.402 & 0.6134 & 2.432 & 3.665 & 5.002 & 7.060 & 1.950 & 1.782 & 1.099 & \\

\bottomrule

\end{tabular}

}
\label{tab:AccLidar}
\end{table*}

\begin{table*}[t]
\centering
\caption{Accuracy Performance on Visual Degradation. Red numbers represent ATE ranking. * denotes incomplete submissions.}
\scalebox{0.75}{
\begin{tabular}{@{}l|cccccc|ccc|c@{}}
\toprule
 \multicolumn{1}{c|}{}& \multicolumn{6}{c|}{\color{Dark} \textbf{Visual Degradation (Real World)}} & \multicolumn{3}{c|}{\color{Dark} \textbf{Simulation}} & \\

\cmidrule(lr{0.75em}){2-7} \cmidrule(lr{0.75em}){8-10} 

\multicolumn{1}{c|}{\textbf{Team}} & \color{Dark} Lowlight 1 & \color{Dark} Lowlight 2& \color{Dark} Over Exposure   & \color{Dark} Flash Light  & \color{Dark} Smoke Room  & \color{Dark} Outdoor Night  &\color{Dark} End of World  & \color{Dark} Moon & \color{Dark} Western Desert   &  \textbf{Average}     \\ 

\cmidrule{1-2}  \cmidrule(lr{0.75em}){3-3} \cmidrule(lr{0.75em}){4-4} \cmidrule(lr{0.75em}){5-5} \cmidrule(lr{0.75em}){6-6} \cmidrule(lr{0.75em}){7-7} \cmidrule(lr{0.75em}){8-8} \cmidrule(lr{0.75em}){9-9} \cmidrule(lr{0.75em}){10-10} \cmidrule(l{0.75em}){11-11} 

Peng  et al\textsuperscript{\color{Red}\textbf{1}} & 1.063 &1.637&  \color{UnderWaterDark}{\textbf{0.503}} & \color{UnderWaterDark}{\textbf{0.44}} & \color{UnderWaterDark}{\textbf{0.153}} & \color{UnderWaterDark}{\textbf{0.827}} & \color{UnderWaterDark}{\textbf{0.038}} & \color{UnderWaterDark}{\textbf{0.195}} & \color{UnderWaterDark}{\textbf{0.070}} & \color{UnderWaterDark}{\textbf{0.547}} \\

Thien et al\textsuperscript{\color{Red}\textbf{2}}& 1.081 & 2.054 & 1.733 & 1.054 & 10.532 & 7.692 & 0.753 & 1.228 & 1.209 & 3.037 \\

Jiang et al\textsuperscript{\color{Red}\textbf{3}} & \color{UnderWaterDark}{\textbf{1.019}} &  \color{UnderWaterDark}{\textbf{1.126}} &1.911 & 2.341 & 3.757 & 11.821 & 2.154 & 0.604 & 4.010 & 3.193 
\\

Li et al\textsuperscript{\color{Red}\textbf{4}}  & 5.768 & 7.834 & 1.757 & 1.295 & 5.370 & 10.766 & - & 30.07 & - & 8.98\textsuperscript{\textbf{*}}\\
\bottomrule

\multicolumn{1}{c|}{\textbf{Average}}  & 2.232 & 3.163 & 1.476 & 1.282 & 4.953 & 7.776 & 0.982 & 8.024 & 1.763 &  \\

\bottomrule


\end{tabular}
}
\label{tab:AccVisual}
\end{table*}

\begin{figure*}[t]
    \centering
    \includegraphics[width=0.85\textwidth]{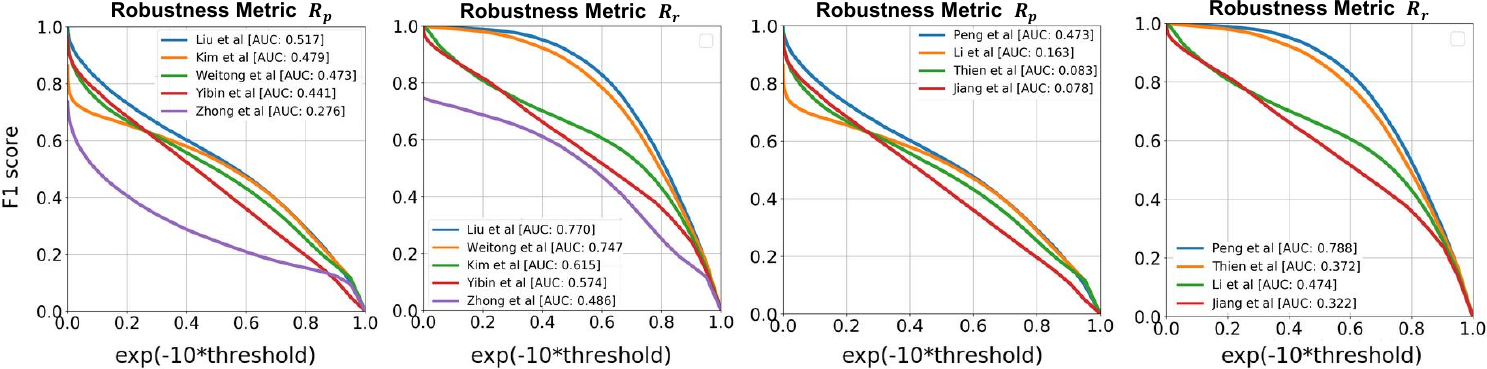}
    \caption{From left to right, it shows robustness metric $R_p$ and $R_r$ for LiDAR and visual sequences respectively. Note: This is a summary of results for all sequences, with weights based on the trajectory length. The area under the curve (AUC) represents the robustness ($R_p, R_r $). The x-axis shows velocity thresholds for classifying estimated velocities as inliers and the y-axis is F-1 score.} 
    \label{fig:robustness_metric}
    \vspace{-0.3cm}
\end{figure*}

\subsection{Ground Truth}\label{sec:ground_truth}

\paragraph{Ground Truth Map}
As depicted in \fref{fig:groundtruth_map}, we utilized the high-precision FARO Focus 3D S120 3D scanner for creating ground-truth maps. This cutting-edge 3D scanner can measure distances up to 120m, with a maximum measurement rate of 976K points per second. The accuracy of the ground truth map is noteworthy, maintaining a range error within $\pm2$mm. We have developed ground truth models for diverse environments, including subterranean areas like urban areas (350m$\times$350m), caves (150m$\times$200m), and tunnels (100$\times$200m), as well as indoor-outdoor spaces (200m$\times$200m) within the CMU campus. To reduce drift, a loop closure algorithm was applied to correct the poses. Notably, 96\% of the scans achieved a position uncertainty of less than 2mm. This ground-truth map is crucial for SLAM development where precision is of utmost importance.

\paragraph{Ground Truth Trajectory}
Ground truth trajectories for all sequences are generated based on the ground truth map.  
In this process, we establish point-to-point, point-to-plane, and point-to-line correspondences\cite{segal2009generalized} between the Ground Truth map and the current LiDAR scan.
Our estimated ground truth trajectory is a comprehensive fusion of various constraints derived from the ground truth map, visual odometry, LiDAR odometry, and IMU measurements. This approach ensures the robustness of our solution across diverse environmental conditions.
To verify accuracy, we reconstruct LiDAR maps from ground truth trajectory and compare it with our ground truth maps.

\subsection{Robustness Metric}
\label{sec:robustness}

As dicussed in \sref{sec:intro}, existing evaluation metrics such as absolute trajectory error (ATE) \cite{Burri25012016} have limitations in evaluating the SLAM's robustness in real-world applications. 
ATE primarily focuses on the accuracy of trajectory and does not consider the completeness of trajectory (recall). Also, it can not effectively capture velocity changes that have a direct impact on the robot's safety. For example,  a pose spike error might not be immediately noticeable in ATE evaluation, it may result in a crash in an aerial robot control.
To address these gaps, we introduce a new \textit{robustness} metric based on estimated velocity. That is because velocity estimation from SLAM is crucial for the robot's control system which directly impacts stability, affecting the robot's safety.
The new robustness metric is the area under the curve (AUC) of the F-1 score:
\begin{equation}
F_1(e) = \frac{2P(e<{T})R(e<{T})}{P(e<{T}) + R(e<{T})},
\end{equation}
where the precision $P$ quantifies the precision of the estimated velocity as a percentage: how closely the estimated velocity points lie to the ground truth point and the recall rate $R$ quantifies the estimated velocity's completeness: to what extent all the ground-truth points are covered. A high F-1 score can only be achieved by the velocity estimation that is both accurate and complete throughout the entire run.
Specifically, an estimated error $e$ is regarded as an inlier,  if it is smaller than a threshold ${T}$. To scale the threshold ${T}$ within the range of $[0, 1]$, we apply exponential mapping $\exp(-10T)$ when calculating robustness metric.

\paragraph{Position Robustness $R_{p}$ \& Rotation Robustness $R_{r}$}
The AUC of F-1 score can be defined using both linear velocity and angular velocity, which can reflect the robustness of position and rotation estimation, respectively. Specifically, we define $R_p = \rm{AUC}(F_1(v_e))$ and $R_{r} = \rm{AUC}(F_1(\omega_e))$, where $v_e$ and $\omega_e$ are the estimated error of linear velocity and rotation velocity, respectively. Note that not all SLAM solutions can output the velocity at the desired frequency. To address this, we use B-splines \cite{mueggler2018continuous,pypose} to obtain smooth trajectories, compute derivatives for smooth trajectories, and derive estimated linear velocity and angular velocity. For more details, please refer to the supplementary.

\section{Open SLAM Challenge Results}
\label{sec:experiments}

In this section, we will present our works from two perspectives: accuracy evaluation and robustness evaluation. 
      
\subsection{Accuracy Evaluation}
\label{sec:accuracy_exp}

The results of the ICCV SLAM challenge underscore the necessity for advancements in system robustness. From 29 submissions, we identified 5 winners in the LiDAR category and 4 in the visual category. However, in the sensor fusion track, which addresses both visual and geometric degradation, no submissions met the criteria for success. This result reveals the existing SLAM systems still have lots of space to improve. Sequence characteristics will be detailed in the supplementary material.
\tref{tab:iccv23} shows the results for both the LiDAR and visual tracks. Unfortunately, there are no current solutions that can balance high accuracy and real-time performance in challenging environments. Since several environments lack geometric features or visual features, the algorithms required more processing times to achieve reasonably high accuracy.
Liu et al. at the University of Hong Kong, who achieved the highest accuracy, recorded an Absolute Trajectory Error (ATE) of 0.588m, but each iteration took 51.3 seconds. In the visual track, Peng et al. from the Samsung Research Center attained a leading ATE of 0.547m, but each iteration required 183 seconds.

\begin{table*}[t]
\centering
\caption{Robustness Performance on Geometric Degradation. Red numbers are robustness ranking. Larger values indicate better robustness.}
\vspace{-3pt}
\scalebox{0.68}{
\begin{tabular}{@{}l|c|cccccc|ccc|ccc|c@{}}
\toprule

\multicolumn{2}{c|}{} &\multicolumn{6}{c|}{\color{Dark} \textbf{Geometric Degradation (Real World)}} &\multicolumn{3}{c|}{\color{Dark} \textbf{Simulation}} &\multicolumn{3}{c|}{\color{Dark} \textbf{Mix Degradation}}&\\

\cmidrule(lr{0.75em}){3-8}  \cmidrule(lr{0.75em}){9-11} \cmidrule(lr{0.75em}){12-14} 
 
\multicolumn{2}{l|}{\textbf{Team}} & \color{Dark} Urban  & \color{Dark} Tunnel & \color{Dark} Cave & \color{Dark} Nuclear\_1  & \color{Dark} Nuclear\_2  & \color{Dark} {Laurel\_Caverns}   &  \color{Dark}Factory &  \color{Dark}Ocean &   \color{Dark}Sewerage & \color{Dark}Long Corridor & \color{Dark}Multi Floor & \color{Dark}Block Lidar & \textbf{Average}\\ 

\cmidrule{1-2}  \cmidrule(lr{0.75em}){3-3} \cmidrule(lr{0.75em}){4-4} \cmidrule(lr{0.75em}){5-5} \cmidrule(lr{0.75em}){6-6} \cmidrule(lr{0.75em}){7-7} \cmidrule(lr{0.75em}){8-8} \cmidrule(lr{0.75em}){9-9} \cmidrule(lr{0.75em}){10-10} \cmidrule(l{0.75em}){11-11} \cmidrule(lr{0.75em}){12-12} \cmidrule(lr{0.75em}){13-13} \cmidrule(lr{0.75em}){14-14} \cmidrule(lr{0.75em}){15-15}

Liu et al\textsuperscript{\color{Red}\textbf{1}}  & \multirow{5}{*}{$R_p\uparrow$}& \color{UnderWaterDark}{\textbf{0.811}}  & \color{UnderWaterDark}{\textbf{0.865}} & 0.736 & 0.747 & 0.504            & \color{UnderWaterDark}{\textbf{0.816}} &  0.157 & 0.135 & \color{UnderWaterDark}{\textbf{0.135}}  & 0.396  & \color{UnderWaterDark}{\textbf{0.529}} & 0.371 & \color{UnderWaterDark}{\textbf{0.516}}    \\

Weitong et al\textsuperscript{\color{Red}\textbf{2}}  &  & 0.773 & 0.838 & 0.690           & \color{UnderWaterDark}{\textbf{0.871}}   & \color{UnderWaterDark}{\textbf{0.739}}   & 0.152      & 0.153 & 0.149 & 0.081     & 0.407  & 0.456 & 0.345 & 0.471  \\

Kim et al\textsuperscript{\color{Red}\textbf{4}} & & 0.747          & \color{UnderWaterDark}{\textbf{0.865}}          & \color{UnderWaterDark}{\textbf{0.737}}            & 0.870          & 0.689          & 0.777 & 0.014         & 0.001         & 0.010          & \color{UnderWaterDark}{\textbf{0.410}}   & 0.285 & 0.345 & 0.479  \\

Yibin et al\textsuperscript{\color{Red}\textbf{3}} & & 0.650          & 0.783          & 0.557    & 0.721  & 0.490 & 0.481 & \color{UnderWaterDark}{\textbf{0.219}}  & \color{UnderWaterDark}{\textbf{0.560}}  & 0.019 & 0.175  & 0.260 & \color{UnderWaterDark}{\textbf{0.370}} & 0.440    \\



Zhong et al\textsuperscript{\color{Red}\textbf{5}} & & 0.567 & 0.683 & 0.204 & 0.680 & 0.426 & 0.447 & 0.103 & 0.084 & 0.122 & 0 & 0 & 0 & 0.276 \\

\cmidrule{1-2}  \cmidrule(lr{0.75em}){3-3} \cmidrule(lr{0.75em}){4-4} \cmidrule(lr{0.75em}){5-5} \cmidrule(lr{0.75em}){6-6} \cmidrule(lr{0.75em}){7-7} \cmidrule(lr{0.75em}){8-8} \cmidrule(lr{0.75em}){9-9} \cmidrule(lr{0.75em}){10-10} \cmidrule(l{0.75em}){11-11} \cmidrule(lr{0.75em}){12-12} \cmidrule(lr{0.75em}){13-13} \cmidrule(lr{0.75em}){14-14} \cmidrule(lr{0.75em}){15-15}

Liu et al\textsuperscript{\color{Red}\textbf{1}}  & \multirow{5}{*}{$R_r\uparrow$}& \color{UnderWaterDark}{\textbf{0.888}}  & \color{UnderWaterDark}{\textbf{0.893}} & \color{UnderWaterDark}{\textbf{0.816}} & 0.857 & 0.778              & \color{UnderWaterDark}{\textbf{0.861}}    &  0.692& 0.719 & \color{UnderWaterDark}{\textbf{0.652}} & 0.753  & 0.689 & \color{UnderWaterDark}{\textbf{0.643}} & \color{UnderWaterDark}{\textbf{0.770}}   \\

Weitong et al\textsuperscript{\color{Red}\textbf{2}}  &  & 0.886 & 0.892 & 0.811& \color{UnderWaterDark}{\textbf{0.893}} & \color{UnderWaterDark}{\textbf{0.840}} & 0.598  & 0.688& 0.720 & 0.514  & 0.766 & \color{UnderWaterDark}{\textbf{0.706}} & 0.644 & 0.746  \\

Kim et al\textsuperscript{\color{Red}\textbf{4}} & & 0.680          & \color{UnderWaterDark}{\textbf{0.893}}         & \color{UnderWaterDark}{\textbf{0.816}}            & \color{UnderWaterDark}{\textbf{0.893}}          & 0.803         & 0.808 & 0.198          & 0.125        & 0.300          & \color{UnderWaterDark}{\textbf{0.802}}   & 0.422 & 0.642 & 0.615  \\

Yibin et al\textsuperscript{\color{Red}\textbf{3}} & & 0.624          & 0.707          & 0.462            & 0.733         & 0.540         & 0.467          & \color{UnderWaterDark}{\textbf{0.862}}          & \color{UnderWaterDark}{\textbf{0.962}}          & 0.496 & 0.477  & 0.342 & 0.213 & 0.573   \\



Zhong et al\textsuperscript{\color{Red}\textbf{5}} & & 0.731 & 0.745 & 0.437 & 0.757 & 0.603 & 0.523 & 0.688 & 0.704 & 0.643 & 0 & 0 & 0 & 0.485 \\
\bottomrule

\multicolumn{2}{c|}{\textbf{Average $R_p$}}  & 0.710 & 0.807 & 0.585 & 0.778 & 0.570 & 0.535 & 0.130 & 0.186 & 0.073 & 0.278 & 0.306 & 0.286 & \\
\multicolumn{2}{c|}{\textbf{Average $R_r$}}  & 0.762 & 0.826 & 0.668 & 0.827 & 0.713 & 0.651 & 0.626 & 0.646 & 0.521 & 0.560 & 0.432 & 0.428 & \\
\bottomrule

\end{tabular}
}
\label{tab:RobustLidar}
\end{table*}

\begin{table*}[t]
\centering
\caption{Robustness Performance on Visual Degradation. Red numbers are robustness ranking. Larger values indicate better robustness.}
\vspace{-3pt}
\scalebox{0.74}{
\begin{tabular}{@{}l|c|cccccc|ccc|c@{}}
\toprule
 \multicolumn{2}{c|}{}& \multicolumn{6}{c|}{\color{Dark} \textbf{Visual Degradation (Real World)}} & \multicolumn{3}{c|}{\color{Dark} \textbf{Simulation}} & \\

\cmidrule(lr{0.75em}){3-8} \cmidrule(lr{0.75em}){9-11} 

\multicolumn{2}{l|}{\textbf{Team}} & \color{Dark} Lowlight 1 & \color{Dark} Lowlight 2& \color{Dark} Over Exposure   & \color{Dark} Flash Light  & \color{Dark} Smoke Room  & \color{Dark} Outdoor Night  &\color{Dark} End of World  & \color{Dark} Moon & \color{Dark} Western Desert   &  \textbf{Average}     \\ 

\cmidrule{1-2}  \cmidrule(lr{0.75em}){3-3} \cmidrule(lr{0.75em}){4-4} \cmidrule(lr{0.75em}){5-5} \cmidrule(lr{0.75em}){6-6} \cmidrule(lr{0.75em}){7-7} \cmidrule(lr{0.75em}){8-8} \cmidrule(lr{0.75em}){9-9} \cmidrule(lr{0.75em}){10-10} \cmidrule(l{0.75em}){11-11} \cmidrule(l{0.75em}){12-12} 

Peng  et al\textsuperscript{\color{Red}\textbf{1}} &\multirow{5}{*}{$R_{p}\uparrow$} & \color{UnderWaterDark}{\textbf{0.357}} & \color{UnderWaterDark}{\textbf{0.227}} & \color{UnderWaterDark}{\textbf{0.264}} & 0.203 & \color{UnderWaterDark}{\textbf{0.536}} & \color{UnderWaterDark}{\textbf{0.270}} & \color{UnderWaterDark}{\textbf{0.699}} & \color{UnderWaterDark}{\textbf{0.893}} & \color{UnderWaterDark}{\textbf{0.806}} & \color{UnderWaterDark}{\textbf{0.472}} \\

Thien et al\textsuperscript{\color{Red}\textbf{3}}& & 0.045 & 0.070 & 0.240 & 0.156 & 0.131 & 0.075 & 0 & 0.031 & 0 & 0.083 
\\

Jiang et al \textsuperscript{\color{Red}\textbf{4}}  & & 0.046 & 0.039 & 0.194 & 0.088 & 0.242   & 0.095 & 0& 0 & 0 & 0.078
\\


Li et al\textsuperscript{\color{Red}\textbf{2}} &  & 0.342 & 0.187 & 0.257 & \color{UnderWaterDark}{\textbf{0.208}} & 0.322 & 0.142 & 0 & 0.006 & 0 & 0.162\\




\cmidrule{1-2}  \cmidrule(lr{0.75em}){3-3} \cmidrule(lr{0.75em}){4-4} \cmidrule(lr{0.75em}){5-5} \cmidrule(lr{0.75em}){6-6} \cmidrule(lr{0.75em}){7-7} \cmidrule(lr{0.75em}){8-8} \cmidrule(lr{0.75em}){9-9} \cmidrule(lr{0.75em}){10-10} \cmidrule(l{0.75em}){11-11} \cmidrule(l{0.75em}){12-12} 

Peng  et al\textsuperscript{\color{Red}\textbf{1}} &\multirow{5}{*}{$R_r\uparrow$} &  \color{UnderWaterDark}{\textbf{0.641}} & \color{UnderWaterDark}{\textbf{0.581}} & \color{UnderWaterDark}{\textbf{0.744}} & 0.650 & \color{UnderWaterDark}{\textbf{0.878}} & \color{UnderWaterDark}{\textbf{0.670}} & \color{UnderWaterDark}{\textbf{0.975}} & 
  \color{UnderWaterDark}{\textbf{0.975}} &  \color{UnderWaterDark}{\textbf{0.974}} &  \color{UnderWaterDark}{\textbf{0.787}}\\

Thien et al\textsuperscript{\color{Red}\textbf{3}}& & 0.413 & 0.445 & 0.610 & 0.269 & 0.474 & 0.487 & 0.177 & 0.315 & 0 &  0.354
\\

Jiang et al\textsuperscript{\color{Red}\textbf{4}}&  & 0.453 & 0.452 & 0.619 & 0.252 & 0.542 & 0.577 & 0.002 & 0 & 0.157 & 0.339 
\\


Li et al\textsuperscript{\color{Red}\textbf{2}} &  & 0.642 & 0.574 & 0.657 &  \color{UnderWaterDark}{\textbf{0.651}} & 0.773 & 0.660 & 0.0 & 0.305 & 0 & 0.473 \\

\bottomrule

\multicolumn{2}{c|}{\textbf{Average $R_p$}}  & 0.198 & 0.131 & 0.239 & 0.164 & 0.308 & 0.146 & 0.175 & 0.232 & 0.202 &  \\
\multicolumn{2}{c|}{\textbf{Average $R_r$}}  & 0.537 & 0.513 & 0.658 & 0.456 & 0.667 & 0.598 & 0.288 & 0.399 & 0.283 &  \\
\bottomrule

\end{tabular}
}
\label{tab:RobustVisual}
\vspace{-0.3cm}
\end{table*}

\paragraph{LiDAR Track Discussion} To enhance robustness in all-weather environments, a pivotal question arises: what are the limits of current LiDAR SLAM algorithms? Table \ref{tab:AccLidar} presents a summary of the ATE/RPE errors observed in real-world geometrically degraded, simulated, and mixed degradation scenarios across the top five teams. 
In real-world geometrically degraded environments, we observed the \textbf{first limitation:} existing LiDAR solutions struggle in confined spaces such as caves. Almost all algorithms perform worse in Cave (Figure \ref{fig:various_terrains}S) and Laureal\_Cavern environments (Figure \ref{fig:various_terrains}N) with average ATE 0.696 and 2.432 meters respectively. 
One reason is that cave environments is the most confined space which may lack sufficient geometric features, compared with other environments like tunnel (Figure \ref{fig:various_terrains}J) and Urban (Figure \ref{fig:various_terrains}B) and Nuclear scenes (Figure \ref{fig:various_terrains}U).     

In simulated environments, we encountered the \textbf{second limitation:} existing methods are tailored for fixed motion patterns like vehicles and struggle with unpredictable motion patterns. Since our simulation sequence is derived from TartanAir \cite{tartanair2020iros}, featuring aggressive and random motion patterns, it might not follow the usual velocity distribution domain and has not been thoroughly tested. 
As shown in \tref{tab:AccLidar} (simulation), it resulted in significantly higher ATE errors (5.239m) compared to real-world's (0.836m).

In mixed degradation environments, the \textbf{third limitation} we found is that existing methods cannot actively select the most informative measurements to adapt to new environments. For example, in the Long Corridor sequence (\fref{fig:various_terrains} A) with low lighting and a featureless environment, the scenario is intricately designed for collaborative usage of LiDAR and Visual sensors. Algorithms are expected to utilize visual information in geometrically degraded environments while disregarding it in darkness. Similarly, the Block LiDAR sequences (\fref{fig:various_terrains} V) aim to simulate sensor drop scenarios frequently encountered in robotic applications. These sequences involve alternating periods of LiDAR and Visual data loss, challenging algorithms to promptly detect sensor failures and switch to the other modality for SLAM. However, the average ATE error (1.61m) in \tref{tab:AccLidar} (mixed degradation) is much higher than in geometric degradation scenarios (0.836m), suggesting the above limitations. Details are in the supplementary.

\paragraph{Visual Track Discussion} 
To improve the robustness, a pivotal question arises: what are the limits of current visual SLAM? To what extent does image quality impact a visual SLAM system?
\tref{tab:AccVisual} shows a summary of ATE errors from real-world and simulation scenarios from the awarded 4 teams. 
In real-world scenarios, \textbf{the first limitation} is the lack of anti-noise capability in current methods, especially in low-quality image settings. We assessed visual odometry accuracy in visually degraded environments, including low lighting, sunlight overexposure, flashing lights, smoke-filled rooms, and nighttime outdoor settings (Figure \ref{fig:various_terrains}). Trajectories in conditions like the Smoke Room and Outdoor Night exhibit significantly higher average ATE errors (4.95m, 7.76m respectively). Smoke and night scenes challenge feature extraction and tracking and introduce sensor noise, emphasizing the need for robust anti-noise algorithms. \textbf{The second limitation} is that existing methods struggle to overcome aggressive motion. Even in simulations with relatively good image quality, most methods still show significant average ATE errors (8.024m) on the Moon sequence (\fref{fig:various_terrains} Y). See more details in supplementary.

\subsection{Robustness Evaluation}\label{sec:robust_exp}
\label{sec:robustness_exp}

The F-1 robustness curve provides a detailed evaluation under different error tolerances, while $R_p$ and $R_r$ provide an overall measure of robustness that is not dependent on a specific decision threshold.
This suggests the flexibility of our new robustness metric.
Figure \ref{fig:robustness_metric} clearly shows the robustness performance of all teams regarding position and rotation, highlighting the differences in robustness between the methods. 
The values of $R_p$ and $R_r$, which are the AUC under F-1 curves, are displayed in the brackets of the legends.
It reveals that Liu et al. and Peng et al. are the most robust solutions for LiDAR and visual tracks, respectively.

\paragraph{Is the robustness metric robust?} 
We did extensive experiments on our robustness metric against diverse sensor degradation settings.
Table \ref{tab:RobustLidar} and \ref{tab:RobustVisual} present a summary of robustness performance in real-world, simulated, and mixed degradation scenarios for both LiDAR and visual sequences. 
In Table \ref{tab:RobustLidar}, we observe that the average values of $R_p$ and $R_r$ for the mixed degradation environment (0.290, 0.473) are significantly lower than geometric degradation environments (0.664, 0.754). This observation suggests that the majority of SLAM algorithms exhibit reduced robustness in mixed degradation environments as compared to geometrically degraded ones. This aligns with our expectations and verifies the effectiveness of the robustness metric.

Our robustness ranking, indicated by small red numbers, differs from those on ATE ranking shown in \tref{tab:AccLidar} and \ref{tab:AccVisual}. This is because our F-1 score-based metrics $ R_p $ and $R_r$ jointly consider precision and recall rate to provide a balanced evaluation of SLAM performance, considering both metrics across the full spectrum of thresholds. In contrast, the ATE focuses solely on precision, neglecting the trajectory's completeness (recall).  We also evaluate our metric on various synthetic trajectories in the supplementary.

\section{Conclusion}
We introduce SubT-MRS, a comprehensive SLAM dataset with various sensor data, locomotion patterns, and over 30 degradation types in simulation and real-world settings to push SLAM towards all-weather environments. Additionally, we introduce a new robustness metric to evaluate the reliability of SLAM systems, enhancing robot safety control. 
29 teams have tested SubT-MRS in the organized SLAM challenge and we expect that it will serve as a critical benchmark for future SLAM development.

{
    \small
    \bibliographystyle{ieeenat_fullname}
    \bibliography{main}
}

\clearpage
\setcounter{page}{1}
\appendix
\maketitlesupplementary



\section{Overview}
We will provide materials from three perspectives: description and analysis of the SubT-MRS dataset, derivation of robustness metric, and more experiment results.           

\section{SubT-MRS Dataset}
In this section, we will provide a more detailed description of our dataset, present key statistics to highlight the challenges inherent in datasets and exhibit a gallery of ground truth 3D models from a range of diverse environments.

\subsection{\textbf{Dataset Description}}

\setlength{\tabcolsep}{3pt}

\begin{table*}[!]
\centering
\caption{Detailed Dataset Information on LiDAR Track (Blue shadings indicate ATE rankings)}
\scalebox{0.55}{
\begin{tabular}{@{}l|c|ccccccc|ccc|cccc|c@{}}
\toprule

\multicolumn{2}{c|}{} &\multicolumn{7}{c}{\color{Dark} \textbf{Degradation Type}} &\multicolumn{3}{c}{\color{Dark} \textbf{Motion Type}} &\multicolumn{4}{c|}{\color{Dark} \textbf{Sensor Used}} & \\

\cmidrule(lr{0.75em}){3-9}  \cmidrule(lr{0.75em}){10-12} \cmidrule(lr{0.75em}){13-16} \cmidrule(lr{0.75em}){13-16} 
 
\multicolumn{2}{l|}{\textbf{Dataset Seq}} & \color{Dark} Low Light  & \color{Dark} Textureless   & \color{Dark} Structureless& \color{Dark} Stairs   & \color{Dark} Smoke/Snow & Aggressive Motion & Repetitive Features  &  \color{Dark}Vehicle Type &  \color{Dark}Max Speed &   \color{Dark}Length (m) & \color{Dark}Fisheye & \color{Dark}LiDAR & \color{Dark}Thermal & \color{Dark}IMU & \textbf{Average ATE}\\ 

\cmidrule{1-2}  \cmidrule(lr{0.75em}){3-3} \cmidrule(lr{0.75em}){4-4} \cmidrule(lr{0.75em}){5-5} \cmidrule(lr{0.75em}){6-6} \cmidrule(lr{0.75em}){7-7} \cmidrule(lr{0.75em}){8-8} \cmidrule(lr{0.75em}){9-9} \cmidrule(lr{0.75em}){10-10} \cmidrule(l{0.75em}){11-11} \cmidrule(lr{0.75em}){12-12} \cmidrule(lr{0.75em}){13-13} \cmidrule(lr{0.75em}){14-14} \cmidrule(lr{0.75em}){15-15} \cmidrule(lr{0.75em}){17-17}

Urban & \multirow{6}{*}{\begin{sideways}Real World\end{sideways}}& \cmark   & \cmark  & \xmark  & \xmark  & \cmark          & \xmark    & \xmark    &  UGV1 & 2 m/s & 441.86 & \xmark & \cmark & \xmark & \cmark & \colorbox{bshade!40}{\makebox[6em][c]{0.633}}\\

Tunnel & & \cmark         & \cmark         & \cmark           & \xmark         & \cmark          & \xmark       & \xmark      & UGV2         & 2 m/s         & 493.67 & \xmark & \cmark & \xmark & \cmark & \colorbox{bshade!10}{\makebox[6em][c]{0.240}}\\

Cave &  & \cmark & \cmark         & \cmark & \xmark       & \xmark      & \xmark     & \cmark      & UGV3  & 2 m/s    & 593.79  & \xmark & \cmark & \xmark & \cmark & \colorbox{bshade!40}{\makebox[6em][c]{0.696}}\\

Nuclear\_1 & & \cmark          & \cmark          & \xmark      & \xmark        &  \cmark    & \xmark         & \xmark  & UGV1          & 2 m/s        & 124.92          & \xmark   & \cmark & \xmark & \cmark  & \colorbox{bshade!20}{\makebox[6em][c]{0.402}}\\

Nuclear\_2 & & \cmark          & \cmark          & \xmark      & \xmark        &  \cmark          & \xmark   & \xmark  & UGV2          & 2m/s        & 1377.37          & \xmark   & \cmark & \xmark & \cmark  &\colorbox{bshade!40}{\makebox[6em][c]{0.613}}\\

Laurel\_Cavern & & \cmark          & \cmark          & \cmark      & \xmark        &  \xmark          & \xmark    & \cmark  & Handheld          & 2 m/s        & 490.46          & \cmark   & \cmark & \cmark & \cmark & \colorbox{bshade!50}{\makebox[6em][c]{2.432}}   \\

\cmidrule{1-2}  \cmidrule(lr{0.75em}){3-3} \cmidrule(lr{0.75em}){4-4} \cmidrule(lr{0.75em}){5-5} \cmidrule(lr{0.75em}){6-6} \cmidrule(lr{0.75em}){7-7} \cmidrule(lr{0.75em}){8-8} \cmidrule(lr{0.75em}){9-9} \cmidrule(lr{0.75em}){10-10} \cmidrule(l{0.75em}){11-11} \cmidrule(lr{0.75em}){12-12} \cmidrule(lr{0.75em}){13-13} \cmidrule(lr{0.75em}){14-14} \cmidrule(lr{0.75em}){15-15}

Factory  & \multirow{3}{*}{\begin{sideways}Sim\end{sideways}}& \cmark  & \cmark & \cmark & \xmark & \cmark & \cmark   & \xmark   & Drone & 360 degree/s & 160.7 & \xmark  & \cmark & \xmark & \cmark &\colorbox{bshade!70}{\makebox[6em][c]{3.665}} \\

Ocean  & & \cmark  & \cmark & \cmark & \xmark & \xmark & \cmark   & \xmark  & Drone & 360 degree/s & 127.5 & \xmark  & \cmark & \xmark & \cmark &\colorbox{bshade!90}{\makebox[6em][c]{5.002}}  \\

Sewerage  &  & \cmark  & \cmark & \cmark & \xmark & \xmark & \cmark   & \xmark  & Drone & 360 degree/s & 131.0 & \xmark  & \cmark & \xmark & \cmark &\colorbox{bshade!110}{\makebox[6em][c]{7.060}}  \\

\cmidrule{1-2}  \cmidrule(lr{0.75em}){3-3} \cmidrule(lr{0.75em}){4-4} \cmidrule(lr{0.75em}){5-5} \cmidrule(lr{0.75em}){6-6} \cmidrule(lr{0.75em}){7-7} \cmidrule(lr{0.75em}){8-8} \cmidrule(lr{0.75em}){9-9} \cmidrule(lr{0.75em}){10-10} \cmidrule(l{0.75em}){11-11} \cmidrule(lr{0.75em}){12-12} \cmidrule(lr{0.75em}){13-13} \cmidrule(lr{0.75em}){14-14} \cmidrule(lr{0.75em}){15-15}

Long Corridor  & \multirow{3}{*}{\begin{sideways}Mix\end{sideways}}& \cmark  & \cmark & \cmark & \xmark & \xmark & \cmark   & \xmark   & RC Car & 4 m/s & 616.45 & \cmark  & \cmark & \xmark & \cmark &\colorbox{bshade!60}{\makebox[6em][c]{1.950}} \\

Multi Floor  & & \cmark  & \cmark & \cmark & \cmark & \xmark & \cmark   & \cmark  & Legged Robot & 2 m/s & 270 & \cmark  & \cmark & \xmark & \cmark &\colorbox{bshade!50}{\makebox[6em][c]{1.782}}  \\

Block LiDAR  &  & \xmark  & \xmark & \cmark & \xmark & \xmark & \xmark   & \xmark  & Legged Robot & 2 m/s & 307.55 & \cmark  & \cmark & \xmark & \cmark &\colorbox{bshade!45}{\makebox[6em][c]{1.099}}  \\

\bottomrule
\end{tabular}
}
\label{tab:LidarInfo}
\end{table*}

\setlength{\tabcolsep}{3pt}

\begin{table*}[!h]
\centering
\caption{Detailed Dataset Information on Visual Track (Blue shadings indicate ATE rankings)}
\scalebox{0.54}{
\begin{tabular}{@{}l|c|ccccccc|ccc|cccc|c@{}}
\toprule

\multicolumn{2}{c|}{} &\multicolumn{7}{c}{\color{Dark} \textbf{Degradation Type}} &\multicolumn{3}{c}{\color{Dark} \textbf{Motion Type}} &\multicolumn{4}{c|}{\color{Dark} \textbf{Sensor Used}} & \\

\cmidrule(lr{0.75em}){3-9}  \cmidrule(lr{0.75em}){10-12} \cmidrule(lr{0.75em}){13-16} \cmidrule(lr{0.75em}){13-16} 
 
\multicolumn{2}{l|}{\textbf{Dataset Seq}} & \color{Dark} Low Lighting  & \color{Dark} Textureless   & \color{Dark} Over Exposure& \color{Dark} Darkness   & \color{Dark} Smoke & Aggressive Motion & Repetitive Features  &  \color{Dark}Vehicle Type &  \color{Dark}Max Speed &   \color{Dark}Length (m) & \color{Dark}Fisheye & \color{Dark}LiDAR & \color{Dark}Thermal & \color{Dark}IMU & \textbf{Average ATE}\\ 

\cmidrule{1-2}  \cmidrule(lr{0.75em}){3-3} \cmidrule(lr{0.75em}){4-4} \cmidrule(lr{0.75em}){5-5} \cmidrule(lr{0.75em}){6-6} \cmidrule(lr{0.75em}){7-7} \cmidrule(lr{0.75em}){8-8} \cmidrule(lr{0.75em}){9-9} \cmidrule(lr{0.75em}){10-10} \cmidrule(l{0.75em}){11-11} \cmidrule(lr{0.75em}){12-12} \cmidrule(lr{0.75em}){13-13} \cmidrule(lr{0.75em}){14-14} \cmidrule(lr{0.75em}){15-15} \cmidrule(lr{0.75em}){17-17}

Low Light1 & \multirow{6}{*}{\begin{sideways}Real World\end{sideways}}& \cmark   & \cmark  & \cmark  & \xmark  & \xmark          & \xmark    & \cmark    &  Handheld & 2 m/s & 400.61  & \cmark & \xmark & \cmark & \cmark & \colorbox{bshade!70}{\makebox[6em][c]{2.232}}\\

Low Light2 & & \cmark   & \cmark  & \cmark  & \xmark  & \xmark          & \xmark    & \cmark    &  Handheld & 2 m/s & 583.19  & \cmark & \xmark & \cmark & \cmark & \colorbox{bshade!30}{\makebox[6em][c]{0.633}}\\

Over Exposure &  & \xmark & \xmark         & \cmark  & \xmark       & \xmark      & \xmark     & \xmark      & Legged Robot  & 2 m/s    & 456.26  & \cmark & \xmark & \cmark & \cmark & \colorbox{bshade!80}{\makebox[6em][c]{3.163}}\\

Flash Light &  & \cmark & \cmark         & \cmark  & \xmark       & \xmark      & \xmark     & \xmark      & Legged Robot  & 2 m/s    & 147.75  & \cmark & \xmark & \cmark & \cmark & \colorbox{bshade!50}{\makebox[6em][c]{1.476}}\\

Smoke Room & & \cmark          & \cmark          & \xmark      & \cmark        &  \cmark          & \xmark   & \xmark  & RC car          & 2m/s        & 104.84          & \cmark   & \xmark & \cmark & \cmark  &\colorbox{bshade!90}{\makebox[6em][c]{4.953}}\\

Outdoor Night & & \cmark          & \cmark          & \xmark      & \cmark        &  \xmark          & \xmark    & \xmark  & Legged Robot          & 2 m/s        & 254.03          & \cmark   & \xmark & \cmark & \cmark & \colorbox{bshade!100}{\makebox[6em][c]{7.776}}   \\

\cmidrule{1-2}  \cmidrule(lr{0.75em}){3-3} \cmidrule(lr{0.75em}){4-4} \cmidrule(lr{0.75em}){5-5} \cmidrule(lr{0.75em}){6-6} \cmidrule(lr{0.75em}){7-7} \cmidrule(lr{0.75em}){8-8} \cmidrule(lr{0.75em}){9-9} \cmidrule(lr{0.75em}){10-10} \cmidrule(l{0.75em}){11-11} \cmidrule(lr{0.75em}){12-12} \cmidrule(lr{0.75em}){13-13} \cmidrule(lr{0.75em}){14-14} \cmidrule(lr{0.75em}){15-15}

End of World  & \multirow{3}{*}{\begin{sideways}Sim\end{sideways}}& \cmark  & \cmark & \xmark & \xmark & \xmark & \cmark   & \xmark   & Drone & 350 degree/s & 280 & \cmark  & \xmark & \xmark & \cmark &\colorbox{bshade!70}{\makebox[6em][c]{0.982}} \\

Moon  & & \cmark  & \cmark & \xmark & \xmark & \xmark & \cmark   & \cmark  & Drone & 60 degree/s & 850 & \cmark  & \xmark & \xmark & \cmark &\colorbox{bshade!110}{\makebox[6em][c]{8.024}}  \\

Western Desert  & & \cmark  & \cmark & \xmark & \cmark & \xmark & \cmark   & \cmark  & Drone & 80 degree/s & 600 & \cmark  & \xmark & \xmark & \cmark &\colorbox{bshade!60}{\makebox[6em][c]{1.763}}  \\

\bottomrule
\end{tabular}
}
\label{tab:VisualInfo}
\end{table*}

We presented comprehensive details about our datasets, including specific challenges, types of motion, and the sensors used, as shown in \tref{tab:LidarInfo} and \ref{tab:VisualInfo}. The average ATE (Absolute Trajectory Error) is computed based on results of the top 5 teams and blue-shaded area reflects ranking of ATE.

\paragraph{Overview of LiDAR Track in Real-World}  \tref{tab:LidarInfo} provide comprehensive dataset information from the LiDAR track, encompassing geometric degradation in real-world and simulated environments, as well as mixed degradation involving both geometric and visual factors. It's important to note that place recognition in cave settings presents challenges for backend optimization, owing to the similarity and repetitiveness of geometric features. Consequently, this results in an average ATE of 2.432 m, which is the highest.

\begin{figure*}[!h]
    \centering
    \includegraphics[width=1\textwidth,keepaspectratio]{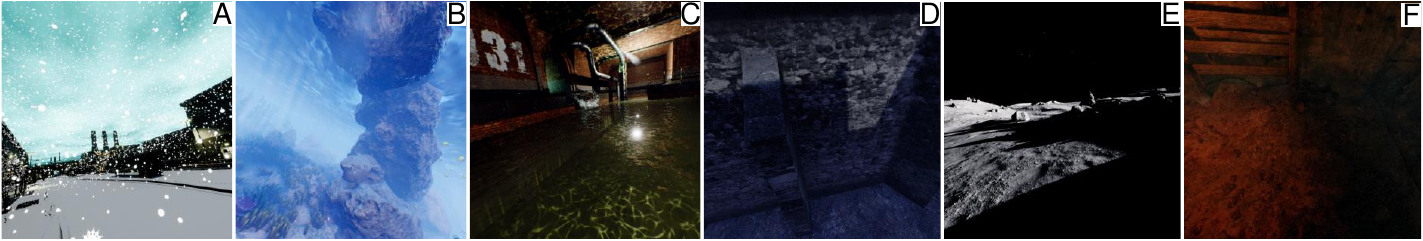}
    \caption{Simulation from both LiDAR/Visual Track. A: Factory B: Ocean  C: Swerage  D: End of world E: Moon F: Western Desert}
    \label{fig:sim_scenes}
\end{figure*}

\paragraph{Overview of LiDAR Track in Simulation}  The dataset was collected using a drone. Its primary challenges arose from the drone's aggressive motion with a maximum angular velocity of 360 degrees per second. Additionally, the Factory sequence, shown in \fref{fig:sim_scenes} A, was captured in snowy conditions. Such a dynamic environment can impair the performance of SLAM systems by introducing snow noise.

\paragraph{Overview of LiDAR Track in Mixed Degradation} It includes both LiDAR and visual degradation such as Long Corridor and Multi-floor sequence, which will be illustrated in Sec \ref{sec:data_analysis}. The Block LiDAR sequence is designed to simulate sensor dropout scenarios, which are frequently encountered in the field of robotics. To evaluate the failure-aware capabilities of the systems, we intentionally disrupted the LiDAR measurements midway through the run. Our objective was to observe whether the algorithm could detect off-normal scenarios and switch to alternate modalities.

\begin{figure}[th]
    \centering
    \includegraphics[width=1.0\linewidth,keepaspectratio]{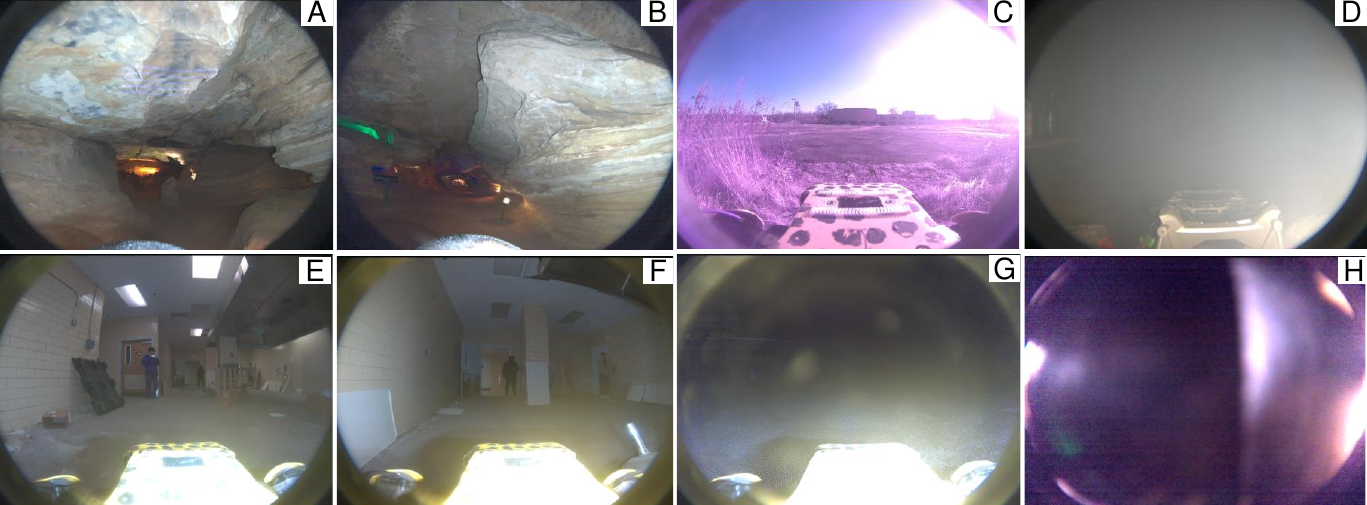}
    \caption{Real-world datasets captured with our fisheye camera to exhibit various visual degradations.}
    \label{fig:challenge_scenes}
\end{figure}

\paragraph{ \textbf{Overview of Visual Track}} Tables \ref{tab:VisualInfo} provide comprehensive information from the visual track, including degradation from real-world and simulated environments. In real-world settings, the Low Light 1 and Low Light 2 sequences, captured in cave environments, presented place recognition challenges due to the similarity and repetitiveness of features, as shown in \fref{fig:challenge_scenes} A and B. These environments look alike despite being in different locations. Figure \ref{fig:challenge_scenes} F, G, and D depict lighting changes from the Flash Light and Over Exposure sequence, which breaks the photometric consistency assumption on feature tracking.
Figure \ref{fig:challenge_scenes} E depicts scenarios from the Smoke Room sequence, where the dynamic smoke not only reduces the number of detected features but also significantly increases noise in feature matching. Figure \ref{fig:challenge_scenes} H exhibits image noise in the Outdoor Night sequence, and I shows the fisheye camera covered to simulate sensor dropout scenarios. In the simulated environments, the scenes captured are shown in Figure \ref{fig:sim_scenes} D, E, and F, respectively.

\begin{figure}[htb]
    \centering
    \includegraphics[width=1.0\linewidth,keepaspectratio]{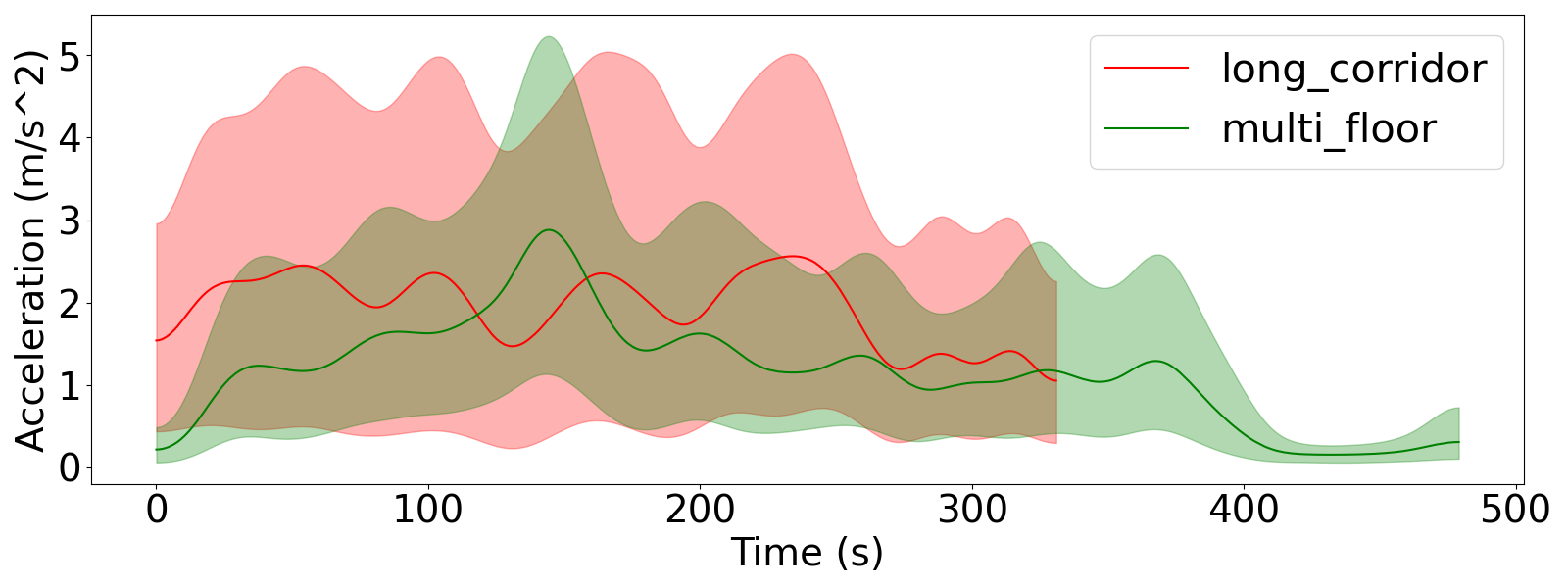}
    \caption{The graphs display average acceleration over time for all runs conducted in the long corridor and multi-floor environments. The shaded area indicates the variance in acceleration.}
    \label{fig:hawkins_accel}
\end{figure}

\begin{figure}[!h]
    \centering
    \includegraphics[width=1.0\linewidth,keepaspectratio]{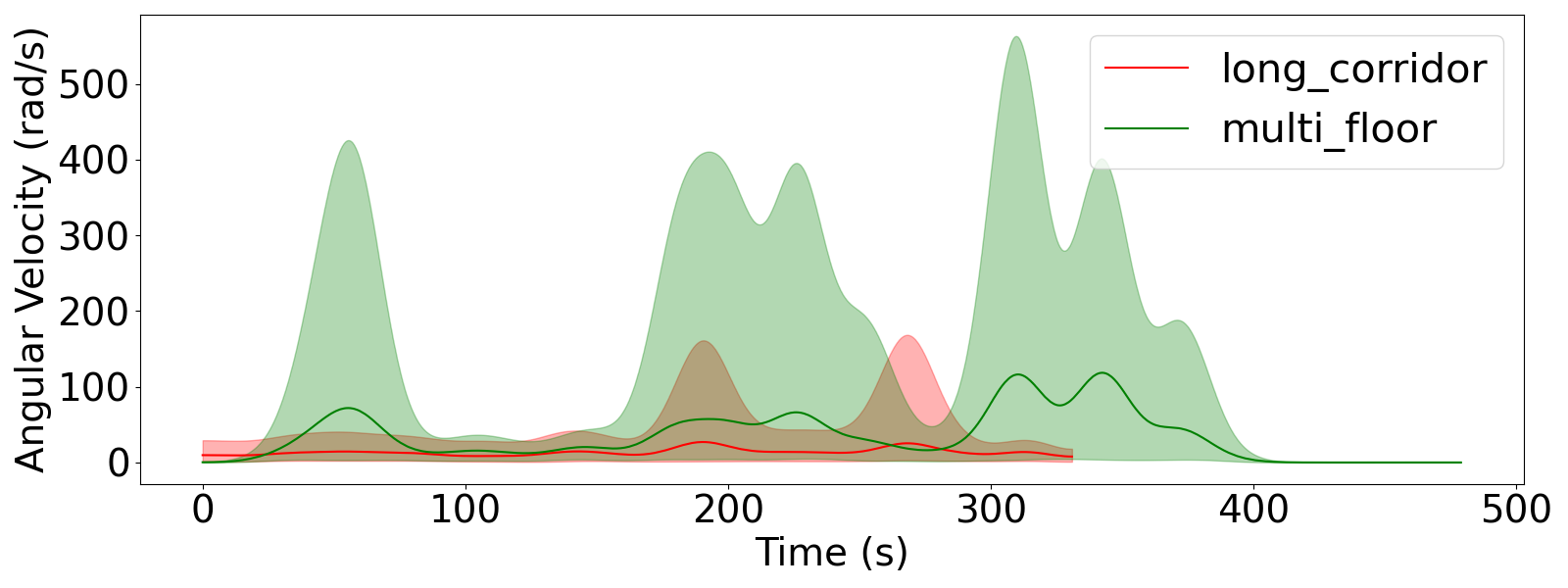}
    \caption{The graphs display average angular velocity over time for all runs conducted in the long corridor and multi-floor environments. The shaded area indicates the variance in angular velocity.}
    \label{fig:hawkins_angular}
\end{figure}

\begin{figure}[htb]
    \centering
    \includegraphics[width=1.0\linewidth,keepaspectratio]{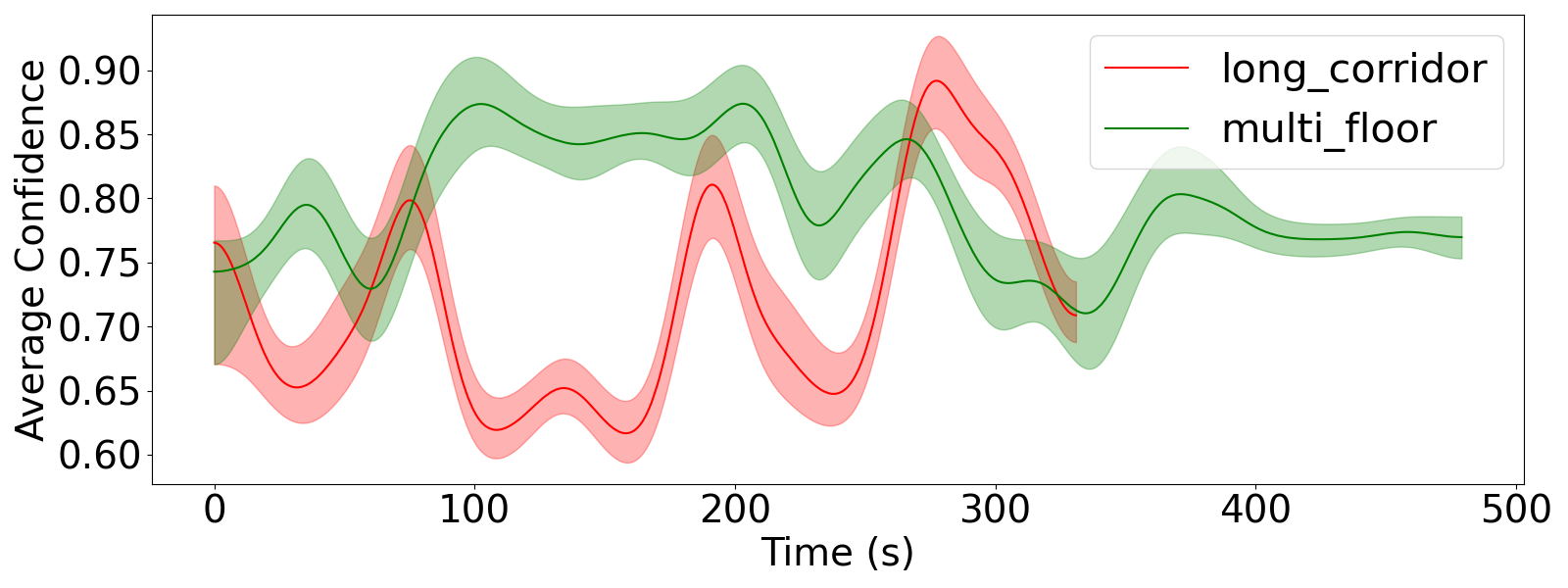}
    \caption{The graphs display the average confidence value of state estimation over time for all runs conducted in the long corridor and multi-floor environments. The shaded area indicates the variance in confidence value.}
    \label{fig:hawkins_confidence}
\end{figure}

\begin{figure}[!h]
    \centering
    \includegraphics[width=1.0\linewidth,keepaspectratio]{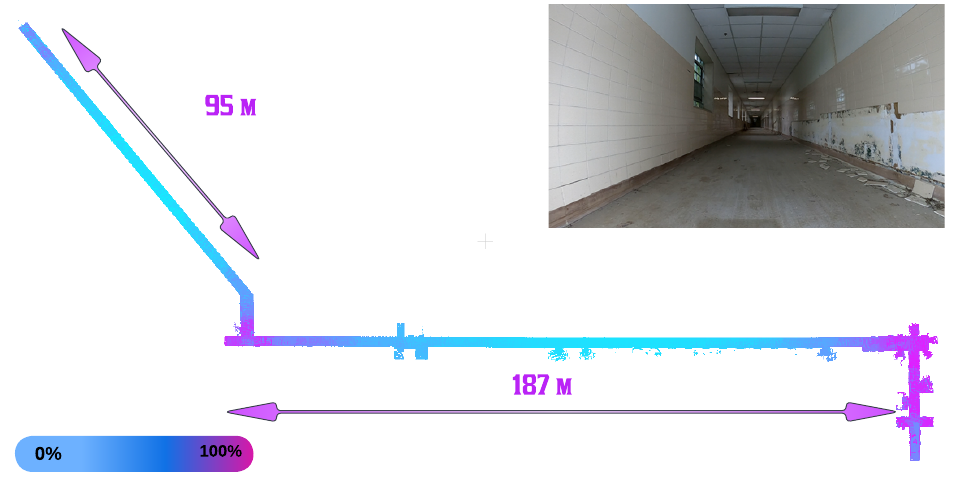}
    \caption{The Confidence Map in Long Corridor Environments. The central region of the long corridor displays the most blue areas, indicating higher uncertainty. This is primarily due to the limited constraints in the forward direction in the corridor, leading to increased mapping uncertainty.}
    \label{fig:uncertainty}
\end{figure}            

\begin{figure*}[!h]
    \centering
    \includegraphics[width=1.0\textwidth,keepaspectratio]{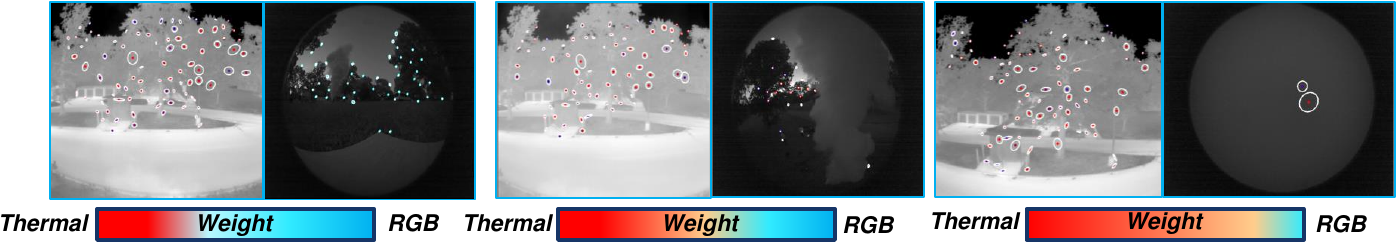}
    \caption{Feature Tracking in Smoke Environments on RGB and Thermal Images: The sequence from left to right on weight illustrates a decrease in feature tracking on RGB images and a corresponding increase on thermal images.}
    \label{fig:visual_degeneracy}
\end{figure*}

\subsection{\textbf{Dataset Statistics}}
\label{sec:data_analysis}

In this section, we will analyze our dataset from three perspectives: degeneracies in LiDAR track, degeneracies in visual track, and an analysis of robot locomotion.

\paragraph{\textbf{Degeneracies in LiDAR}} 
To help users understand the challenges of our dataset, we use various metrics to measure the difficulty of dataset statistically. The statistics include acceleration, angular velocity rate, and confidence value. The acceleration and angular velocity are used to measure the motion pattern of this dataset and a confidence value is provided by Super Odometry\cite{zhao2021super} to evaluate the degradation level of environments. In the LiDAR track, we select 2 typical challenging environments: Long corridor and Multi Floor sequence to evaluate. 

\begin{figure}[ht!]
    \centering
    \includegraphics[width=1.0\linewidth,keepaspectratio]{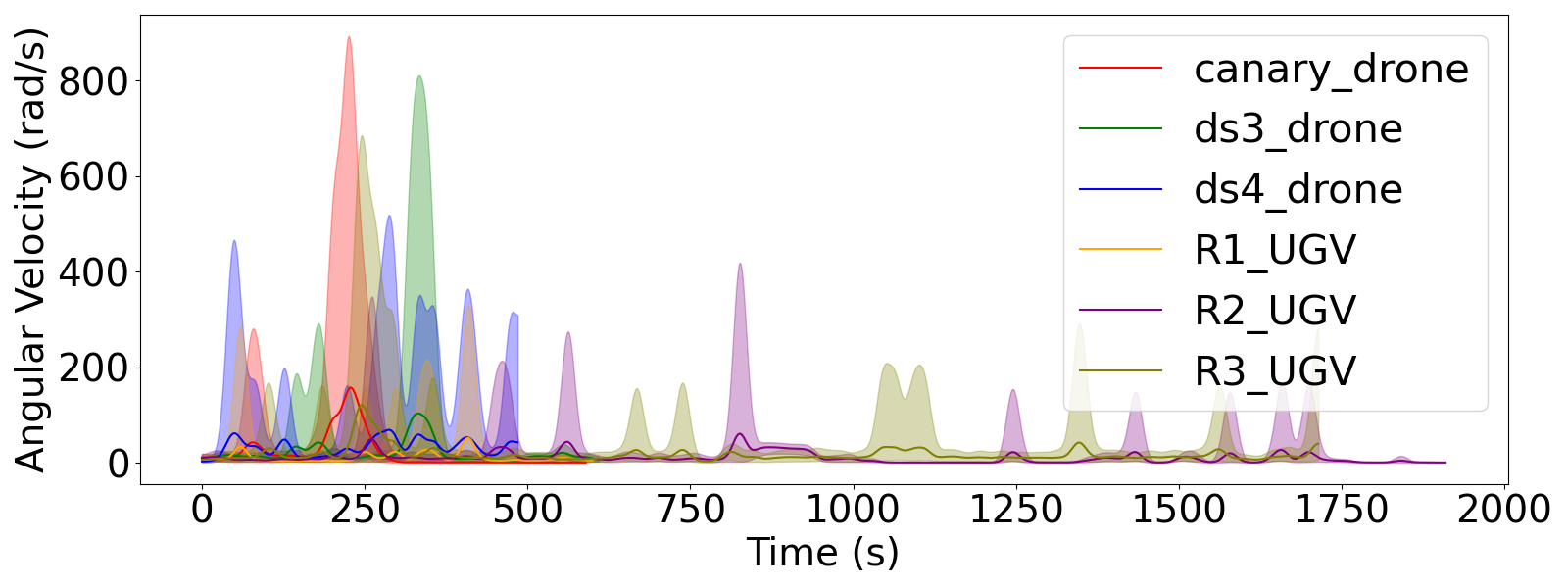}
    \caption{The graph displays the angular velocity over time for all runs conducted in the SubT environment. The shaded area indicates the variance in angular velocity. }
    \label{fig:subt_angular}
\end{figure}

\begin{figure}[ht!]
    \centering
    \includegraphics[width=1.0\linewidth,keepaspectratio]{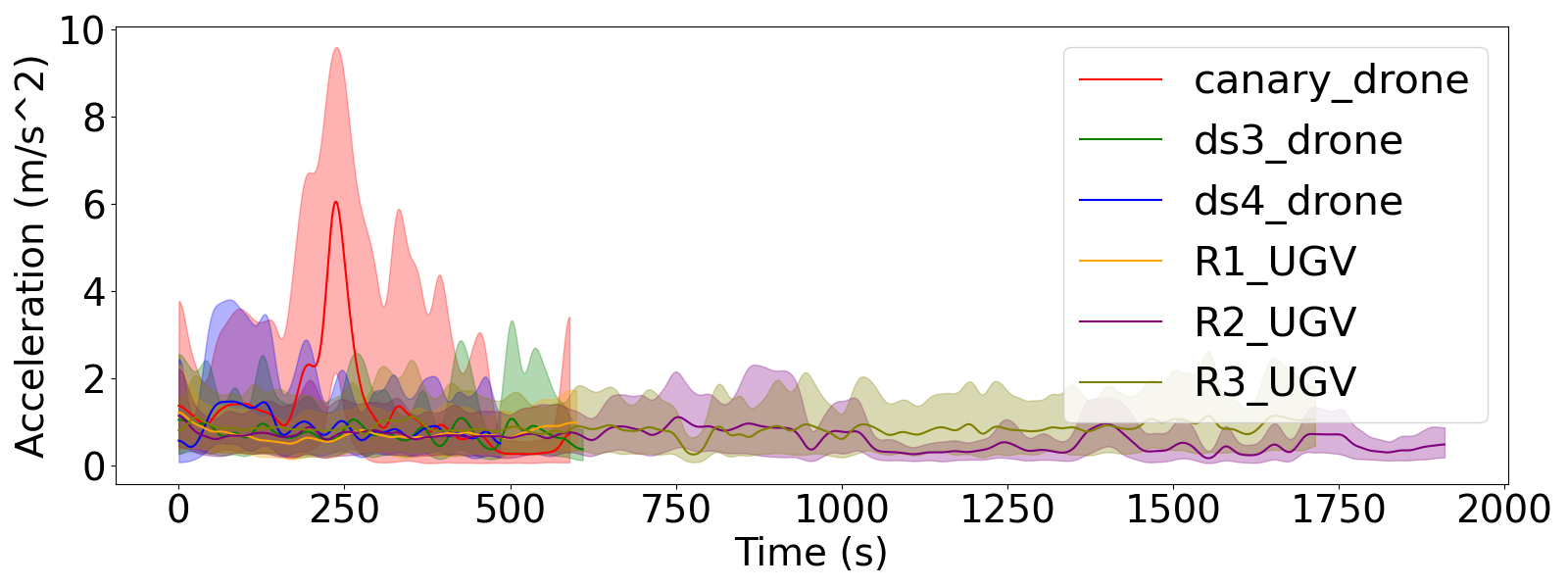}
    \caption{The graph displays the acceleration over time for all runs conducted in the SubT environment. The shaded area indicates the variance in acceleration. }
    \label{fig:subt_accel}
\end{figure}

\fref{fig:subt_accel} and \fref{fig:subt_angular} demonstrate that the Long Corridor sequences exhibit greater acceleration compared to the multi-floor sequences, while the Multi-Floor sequences experience more angular velocity changes than those in long corridor environments.  Given that Super Odometry can assess the confidence of state estimation, we utilize this metric to gauge the overall challenges of the dataset. It is observed that the long corridor shows lower confidence values most of the time, indicating that it is a more challenging environment for LiDAR SLAM systems. \fref{fig:uncertainty} further illustrated the confidence value of map in 3D space.

\paragraph{\textbf{Degeneracies in Visual}}
We selected a smoke environment to specifically illustrate degeneracy, showcasing the tracked feature count in both thermal and RGB images, as shown in \fref{fig:visual_degeneracy}. In this setting, we applied a learning-based feature extraction pipeline\cite{zhao2020tp}, to these multispectral images. With increasing smoke density, the thermal images are not influenced, whereas RGB images tend to add noise to the feature tracking process. Therefore, in such scenarios, an optimal SLAM solution should gradually prioritize the thermal camera as primary sensor for state estimation.

\paragraph{\textbf{Heterogeneous Robot Locomotion }}
Our dataset, sourced from heterogeneous robots, allows for an analysis of their movement patterns to identify which robots yield the more challenging datasets. \fref{fig:subt_accel} and \fref{fig:subt_angular} present the average acceleration and angular velocity across all runs from different robots. These figures highlight that the dataset from the canary drone poses greater challenges compared to others, evidenced by its higher frequency of peaks in both acceleration and angular velocity.


\begin{figure*}[ht]
  \centering
  \begin{subfigure}[b]{1.0\textwidth}
    \centering
    \includegraphics[width=\textwidth]{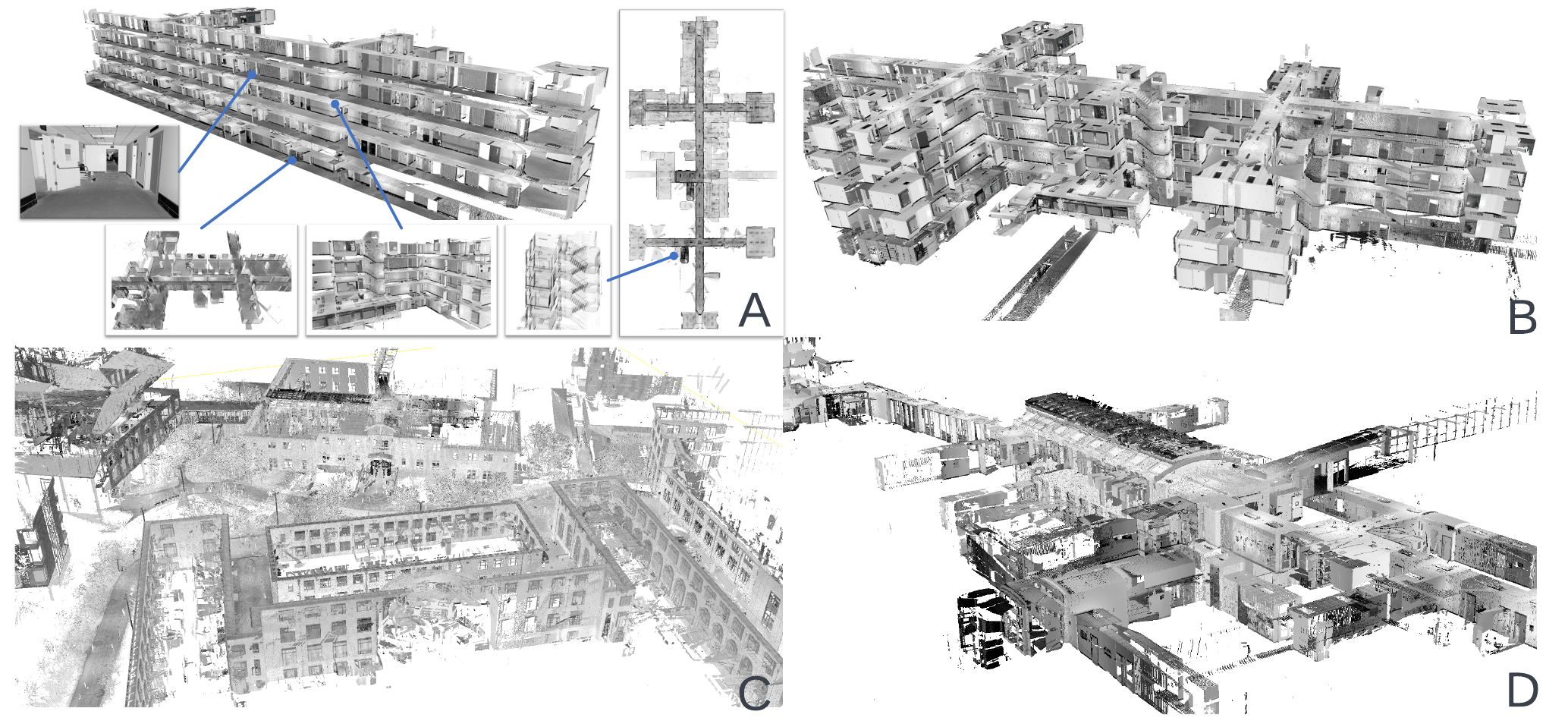}
    \label{fig:dc4}
      \end{subfigure}
 \caption{The Ground Truth Model in SubT-MRS Dataset. A: Long Corridor B: MultiFloor C: Campus Outdoor D: Campus Indoor}
  \label{fig:Hawkins2}
\end{figure*}

\subsection{\textbf{Ground Truth Model}}

In this section, we are pleased to present our extensive collection of Ground Truth 3D Models, as displayed in \fref{fig:Hawkins2}. This collection includes a ground truth model of a university campus, covering both indoor and outdoor areas, extensive loop mapping for long corridor and multi floor environments. Importantly, all ground truth models were captured using a FARO scanner, guaranteeing an accuracy within 10 cm.

\section{Derivation of Robustness Metric}
As introduced in the main paper Sec 3.3, we describe how to use B-spline to derive estimated linear and angular velocity. Additionally, we will show the steps of robustness metric calculation. 

\paragraph{B-splines}
We adopt B-splines \cite{mueggler2018continuous} to interpolate the given trajectory, yielding continuous-time trajectories in $SE(3) $. The continuous trajectory is decided by control poses \(T_{w,i}\), where \(T_{w,i}\) is the estimated poses at time \(t_i\) in a world coordinate system $w$. According to the locality of the cubic B-spline basis, the value of the spline curve at any time \(t\) is decided by four control poses. We use one absolute pose, \(T_{w,i-1}\), and three incremental poses, parametrized by twists $\xi_i$. More specifically, the spline trajectory is given by
\begin{equation}
T_{w,s}(u(t)) = T_{w,i-1} \prod_{j=1}^{3} \exp(\mathbf{B}_j(u(t))\xi_{i+j-1} ) 
\label{eq:1}
\end{equation}
where the $ \exp $ denotes the matrix exponential. \(u(t) = (t - t_i)/\Delta t \in [0, 1]\) and \(t_i = i\Delta t\) are used in the cumulative basis functions for the B-splines. 
\begin{equation}
\bar{\mathbf{B}}(u) = C \begin{bmatrix} 1 \\ u \\ u^2 \\ u^3 \end{bmatrix}, \quad C = \begin{bmatrix} 6 & 0 & 0 & 0 \\ 5 & 3 & -3 & 1 \\ 1 & 3 & 3 & -2 \\ 0 & 0 & 0 & 1 \end{bmatrix},
\end{equation}
In Eq.\ref{eq:1}, \(\mathbf{B}_j\) is the \(j\)-th entry (0-based) of vector \(\mathbf{B}\). The incremental pose from time \(t_{i-1}\) to \(t_i\) is encoded by twist
\begin{equation}
\xi_i = \log(T_{w,i-1}^{-1}T_{w,i}),
\end{equation}
\begin{equation}
\frac{\partial}{\partial t_s} T_{w,s}(u(t)) =[\Delta \dot{R},\Delta \dot{T}]=[\mathbf{\omega}_{w},\mathbf{v}_{w}]
\label{eq:4}
\end{equation}
We differentiate the spline trajectory as presented in Eq.~\ref{eq:4} to obtain estimates of the velocity, denoted as 
$\mathbf{v}_{w}$, and the angular velocity, denoted as $\mathbf{\omega}_{w}$.  We sampled velocities from both the interpolated ground truth trajectory and the trajectory estimated via B-spline, establishing velocity correspondences based on time stamps.

\begin{table*}[t]
\centering
\caption{Accuracy Performance on LiDAR Degradation. Red numbers represent ATE/RPE ranking. * denotes incomplete submissions.}
\scalebox{0.74}{
\begin{tabular}{@{}l|c|cccccc|ccc|ccc|c@{}}
\toprule

\multicolumn{2}{c|}{} &\multicolumn{6}{c}{\color{Dark} \textbf{Real World}} &\multicolumn{3}{c}{\color{Dark} \textbf{Simulation}} &\multicolumn{3}{c}{\color{Dark} \textbf{Mix Degradation}}&\\

\cmidrule(lr{0.75em}){3-8}  \cmidrule(lr{0.75em}){9-11} \cmidrule(lr{0.75em}){12-14} \cmidrule(lr{0.75em}){15-15} 
 
\multicolumn{2}{l|}{\textbf{Team}} & \color{Dark} Urban  & \color{Dark} Tunnel & \color{Dark} Cave & \color{Dark} Nuclear\_1  & \color{Dark} Nuclear\_2  & \color{Dark} {Laurel\_Caverns}   &  \color{Dark}Factory &  \color{Dark}Ocean &   \color{Dark}Sewerage & \color{Dark}Long Corridor & \color{Dark}Multi Floor & \color{Dark}Block Lidar & \textbf{Average}\\ 

\cmidrule{1-2}  \cmidrule(lr{0.75em}){3-3} \cmidrule(lr{0.75em}){4-4} \cmidrule(lr{0.75em}){5-5} \cmidrule(lr{0.75em}){6-6} \cmidrule(lr{0.75em}){7-7} \cmidrule(lr{0.75em}){8-8} \cmidrule(lr{0.75em}){9-9} \cmidrule(lr{0.75em}){10-10} \cmidrule(l{0.75em}){11-11} \cmidrule(lr{0.75em}){12-12} \cmidrule(lr{0.75em}){13-13} \cmidrule(lr{0.75em}){14-14} \cmidrule(lr{0.75em}){15-15}

Liu et al\textsuperscript{\color{Red}\textbf{1}}  & \multirow{5}{*}{\begin{sideways}ATE\end{sideways}}& 0.307  & 0.095 & 0.629 & 0.122 & 0.235              & \color{UnderWaterDark}{\textbf{0.260}}      &    \color{UnderWaterDark}{\textbf{0.889}} & 0.757 & \color{UnderWaterDark}{\textbf{0.978}}  & 1.454 & \color{UnderWaterDark}{\textbf{0.401}} & \color{UnderWaterDark}{\textbf{0.934}}  & \color{UnderWaterDark}{\textbf{0.588}}\\

Yibin et al\textsuperscript{\color{Red}\textbf{4}} & & 1.060          &0.220         & 0.750           & 0.470         & 0.620          & 9.140          & 4.920         & \color{UnderWaterDark}{\textbf{0.280}}         & 24.460  & 2.990 & 5.500 & 1.340 & 4.312 \\

Weitong et al\textsuperscript{\color{Red}\textbf{2}}  &  & \color{UnderWaterDark}{\textbf{0.26}} & 0.096         & \color{UnderWaterDark}{\textbf{0.617}}           & \color{UnderWaterDark}{\textbf{0.120}}        &  \color{UnderWaterDark}{\textbf{0.222}}       & 0.402        & 0.998  & 0.770    & 1.586  & \color{UnderWaterDark}{\textbf{1.254}} & 0.577 & 1.056 & 0.663 \\

Kim et al\textsuperscript{\color{Red}\textbf{3}} & & 0.331          & \color{UnderWaterDark}{\textbf{0.092}}          & 0.787      & 0.123        & 0.270          & 0.279  & 10.628          & 22.425        & 7.147          & 2.100   & 0.650 & 1.068 & 3.825  \\

Zhong et al\textsuperscript{\color{Red}\textbf{5}} & & 1.205 & 0.695 & - & 1.175 & 1.72 & 2.08 & \color{UnderWaterDark}{\textbf{0.889}} & 0.778 & 1.13 & -  & - & - & 1.209\textsuperscript{\textbf{*}}  \\

\cmidrule{1-2}  \cmidrule(lr{0.75em}){3-3} \cmidrule(lr{0.75em}){4-4} \cmidrule(lr{0.75em}){5-5} \cmidrule(lr{0.75em}){6-6} \cmidrule(lr{0.75em}){7-7} \cmidrule(lr{0.75em}){8-8} \cmidrule(lr{0.75em}){9-9} \cmidrule(lr{0.75em}){10-10} \cmidrule(l{0.75em}){11-11} \cmidrule(lr{0.75em}){12-12} \cmidrule(lr{0.75em}){13-13} \cmidrule(lr{0.75em}){14-14} \cmidrule(lr{0.75em}){15-15}

Liu et al\textsuperscript{\color{Red}\textbf{1}}  & \multirow{5}{*}{\begin{sideways}RPE\end{sideways}}& 0.038  & \color{UnderWaterDark}{\textbf{0.032}} & \color{UnderWaterDark}{\textbf{0.055}} & \color{UnderWaterDark}{\textbf{0.028}} & \color{UnderWaterDark}{\textbf{0.048}} & \color{UnderWaterDark}{\textbf{0.040}}  & 0.191 & 0.174 & 0.188 & 0.088  & 0.059 & \color{UnderWaterDark}{\textbf{0.148}} & \color{UnderWaterDark}{\textbf{0.091}}  \\

Yibin et al\textsuperscript{\color{Red}\textbf{3}} & & 0.130          & 0.090          & 0.150            & 0.130          & 0.200          & 0.200          & \color{UnderWaterDark}{\textbf{0.040}}         & \color{UnderWaterDark}{\textbf{0.040}}          & \color{UnderWaterDark}{\textbf{0.160}} & 0.630  & 0.280 & 0.180 & 0.186    \\

Weitong et al\textsuperscript{\color{Red}\textbf{2}}  &  & \color{UnderWaterDark}{\textbf{0.038}} & \color{UnderWaterDark}{\textbf{0.032}}          & 0.056            & 0.029          & 0.049         & 0.046          & 0.190  & 0.183 & 0.243  & \color{UnderWaterDark}{\textbf{0.086}}  & \color{UnderWaterDark}{\textbf{0.054}} & 0.166 & 0.097  \\

Kim et al\textsuperscript{\color{Red}\textbf{4}} & & 0.098         & \color{UnderWaterDark}{\textbf{0.032}}          & 0.055           & \color{UnderWaterDark}{\textbf{0.028}}          & 0.054          & 0.046 & 0.861         & 0.535         & 0.401         & 0.093   & 0.26 & 0.167 & 0.219  \\

Zhong et al\textsuperscript{\color{Red}\textbf{5}} & & 0.157 & 0.062 & - & 0.079 & 0.1062 & 0.0937 & 0.706 & 0.691 & 0.617 & - & - & - & 0.313\textsuperscript{\textbf{*}}  \\
\bottomrule
\end{tabular}
}
\label{tab:ATE_RPE_LIDAR}
\end{table*}

\begin{table*}[t]
\centering
\caption{Accuracy Performance on Visual Degradation. Red numbers represent ATE/RPE ranking. * denotes incomplete submissions.}
\scalebox{0.8}{
\begin{tabular}{@{}l|c|cccccc|ccc|c@{}}
\toprule
 \multicolumn{2}{c|}{}& \multicolumn{6}{c}{\color{Dark} \textbf{Real World}} & \multicolumn{3}{c}{\color{Dark} \textbf{Simulation}} & \\

\cmidrule(lr{0.75em}){3-8} \cmidrule(lr{0.75em}){9-11} \cmidrule(lr{0.75em}){12-12}

\multicolumn{2}{l|}{\textbf{Team}} & \color{Dark} Lowlight 1 & \color{Dark} Lowlight 2& \color{Dark} Over Exposure   & \color{Dark} Flash Light  & \color{Dark} Smoke Room  & \color{Dark} Outdoor Night  &\color{Dark} End of World  & \color{Dark} Moon & \color{Dark} Western Desert   &  \textbf{Average}     \\ 

\cmidrule{1-2}  \cmidrule(lr{0.75em}){3-3} \cmidrule(lr{0.75em}){4-4} \cmidrule(lr{0.75em}){5-5} \cmidrule(lr{0.75em}){6-6} \cmidrule(lr{0.75em}){7-7} \cmidrule(lr{0.75em}){8-8} \cmidrule(lr{0.75em}){9-9} \cmidrule(lr{0.75em}){10-10} \cmidrule(l{0.75em}){11-11} \cmidrule(l{0.75em}){12-12} 

Peng  et al\textsuperscript{\color{Red}\textbf{1}} &\multirow{5}{*}{\begin{sideways}ATE\end{sideways}} & 1.063 &1.637&  \color{UnderWaterDark}{\textbf{0.503}} & \color{UnderWaterDark}{\textbf{0.44}} & \color{UnderWaterDark}{\textbf{0.153}} & \color{UnderWaterDark}{\textbf{0.827}} & \color{UnderWaterDark}{\textbf{0.038}} & \color{UnderWaterDark}{\textbf{0.195}} & \color{UnderWaterDark}{\textbf{0.070}} & \color{UnderWaterDark}{\textbf{0.547}} \\

Jiang et al\textsuperscript{\color{Red}\textbf{3}}&  & \color{UnderWaterDark}{\textbf{1.019}} &  \color{UnderWaterDark}{\textbf{1.126}} &1.911 & 2.341 & 3.757 & 11.821 & 2.154 & 0.604 & 4.010 & 3.193 
\\

Thien et al\textsuperscript{\color{Red}\textbf{2}}& & 1.081 & 2.054 & 1.733 & 1.054 & 10.532 & 7.692 & 0.753 & 1.228 & 1.209 & 3.037 \\

Li et al\textsuperscript{\color{Red}\textbf{4}} &  & 5.768 & 7.834 & 1.757 & 1.295 & 5.370 & 10.766 & - & 30.07 & - & 8.98\textsuperscript{\textbf{*}}\\

\cmidrule{1-2}  \cmidrule(lr{0.75em}){3-3} \cmidrule(lr{0.75em}){4-4} \cmidrule(lr{0.75em}){5-5} \cmidrule(lr{0.75em}){6-6} \cmidrule(lr{0.75em}){7-7} \cmidrule(lr{0.75em}){8-8} \cmidrule(lr{0.75em}){9-9} \cmidrule(lr{0.75em}){10-10} \cmidrule(l{0.75em}){11-11} \cmidrule(l{0.75em}){12-12} 

Peng  et al\textsuperscript{\color{Red}\textbf{1}} &\multirow{5}{*}{\begin{sideways}RPE\end{sideways}} & \color{UnderWaterDark}{\textbf{0.058}} & \color{UnderWaterDark}{\textbf{0.063}} & \color{UnderWaterDark}{\textbf{0.051}} & \color{UnderWaterDark}{\textbf{0.149}} & \color{UnderWaterDark}{\textbf{0.026}} & \color{UnderWaterDark}{\textbf{0.064}} & \color{UnderWaterDark}{\textbf{0.002}} & \color{UnderWaterDark}{\textbf{0.014}} & \color{UnderWaterDark}{\textbf{0.01}} & \color{UnderWaterDark}{\textbf{0.048}} \\

Jiang et al\textsuperscript{\color{Red}\textbf{3}}&  & 0.190 & 0.203 & 0.258 & 0.307 & 0.884 & 3.427 & 3.982 & 0.792 & 7.477 & 1.947 
\\

Thien et al\textsuperscript{\color{Red}\textbf{2}}& & 0.197 & 0.186 & 0.181 & 0.231 & 0.071 & 0.279 & 0.471 & 0.007 & 0.777 &  0.266
\\

Li et al\textsuperscript{\color{Red}\textbf{4}} &  & 0.088 & 0.088 & 0.124 & 0.160 & 0.911 & 0.478 & - & 0.347 & - & 0.314\textsuperscript{\textbf{*}} \\

\bottomrule
\end{tabular}
}
\label{tab:ATE_RPE_VISUAL}
\end{table*}

\paragraph{Robustness Metric Calculation}
We introduce the F-score \cite{knapitsch2017tanks} to evaluate the robustness of SLAM algorithms. The robustness metric can be calculated in the following steps:

\begin{algorithm}
\caption{Robustness Metric Calculation}
\begin{algorithmic}
\State \textbf{Input:} Trajectory $T_{w,i}$, ground truth velocities $\mathbf{v}_{gt}^{w}, \boldsymbol{\omega}_{gt}^{w}$
\State \textbf{Output:} Robustness Metric
\State \textbf{Step 1:} Smooth trajectory with B-spline on $T_{w,i}$
\State \textbf{Step 2:} Derive $\mathbf{v}_{est}^{w}, \mathbf{\omega}_{est}^{w}$ from smoothed trajectory
\State \textbf{Step 3:} Calculate distances:
       $\mathbf{v}_{dis} = ||\mathbf{v}_{gt}^{w} - \mathbf{v}_{est}^{w}||$, 
       $\boldsymbol{\omega}_{dis} = ||\boldsymbol{\omega}_{gt}^{w} - \boldsymbol{\omega}_{est}^{w}||$
\State \textbf{Step 4:} Compute F-scores $F(\mathbf{v}_{dis}), F(\boldsymbol{\omega}_{dis})$ for various thresholds
\State \textbf{Step 5:} Plot F-scores against thresholds; area under curve is robustness metric
\end{algorithmic}
\label{alg:robustness_metric}
\end{algorithm}

\section{Experiments}
In this section, we provide detailed experiments assessing accuracy and robustness, which were not included in the main paper due to length constraints. It should be noted that the main paper already includes comprehensive experiments on our datasets and presents summarized conclusions. The additional experiments here serve as \textbf{supplementary evidence supporting the conclusions} drawn in Sections 4.1 and 4.2 of our paper.

\subsection{Accuracy Evaluation}
In section 4.1, we presented Absolute Trajectory Error (ATE) values for the LiDAR and visual tracks. We omitted the RPE results due to the constraints of paper length.  \tref{tab:ATE_RPE_LIDAR} and \tref{tab:ATE_RPE_VISUAL} provide RPE results. 

\begin{figure*}[t]
    \centering
    \includegraphics[width=0.9\linewidth,keepaspectratio]{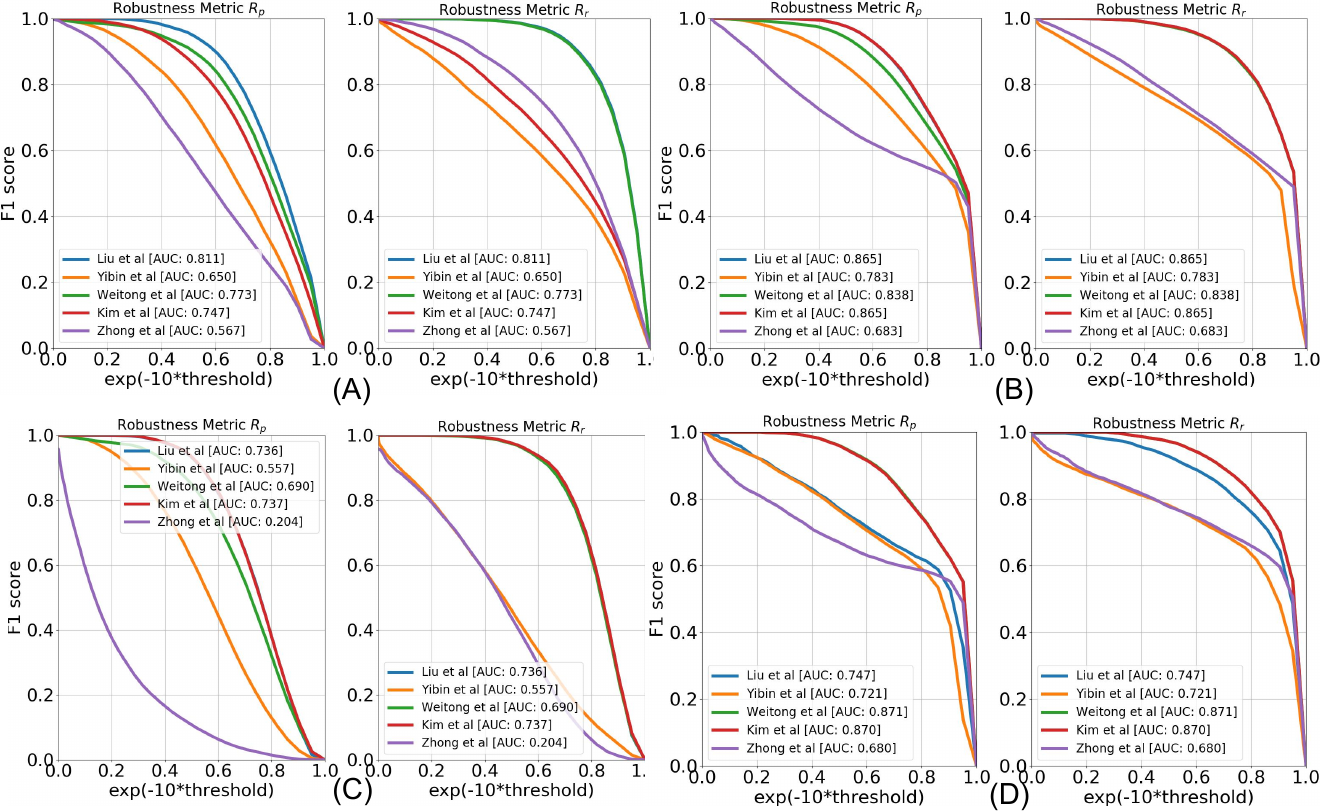}
\includegraphics[width=0.91\linewidth,keepaspectratio]{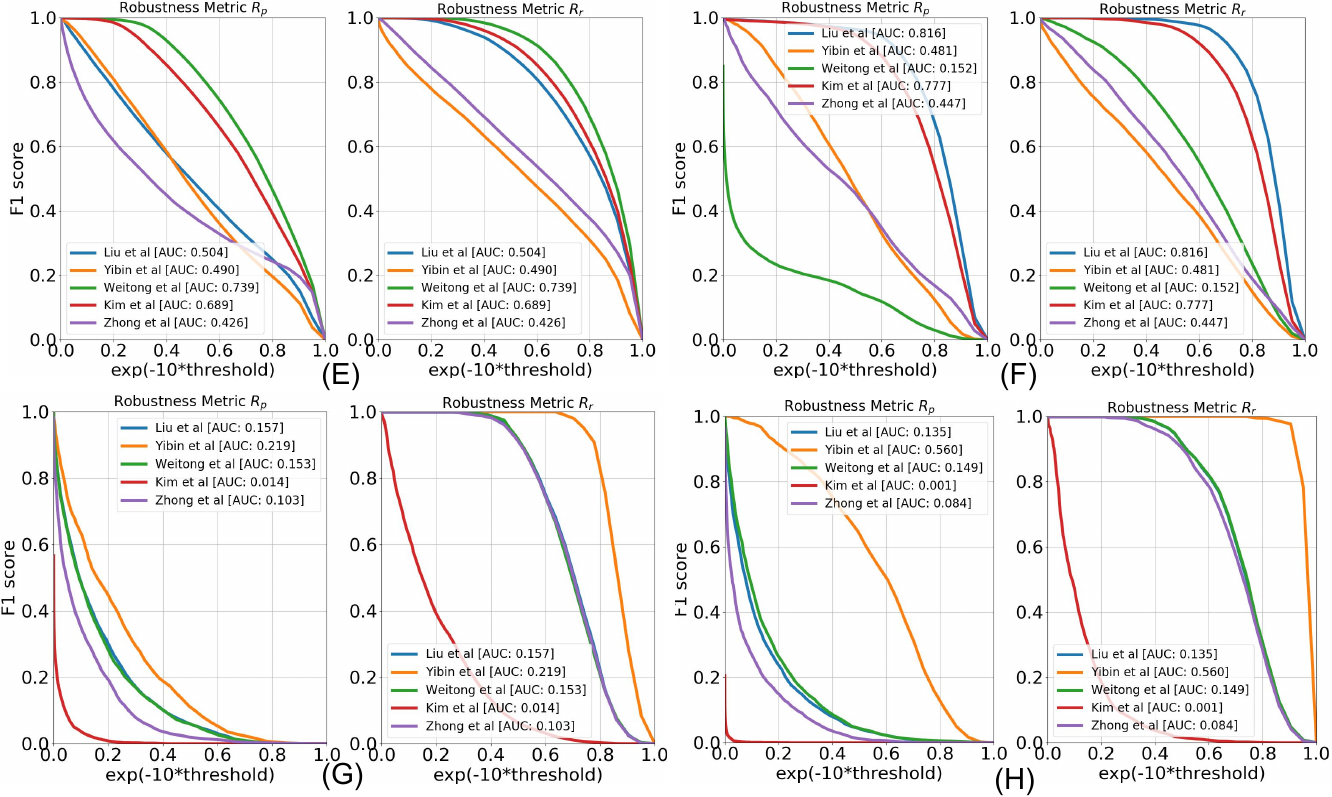}
    \caption{The image, arranged from left to right, displays the robustness metrics  $R_p$ and $R_r$ from LiDAR track teams: (A) Urban, (B) Tunnel, (C) Cave, and (D) Nuclear\_1 (E) Nuclear\_2 (F) Laurel Cavern (G) Factory (H) Ocean}
    \label{fig:robust_lidar1}
\end{figure*}

\begin{figure*}[!h]
    \centering
    \includegraphics[width=0.9\linewidth,keepaspectratio]{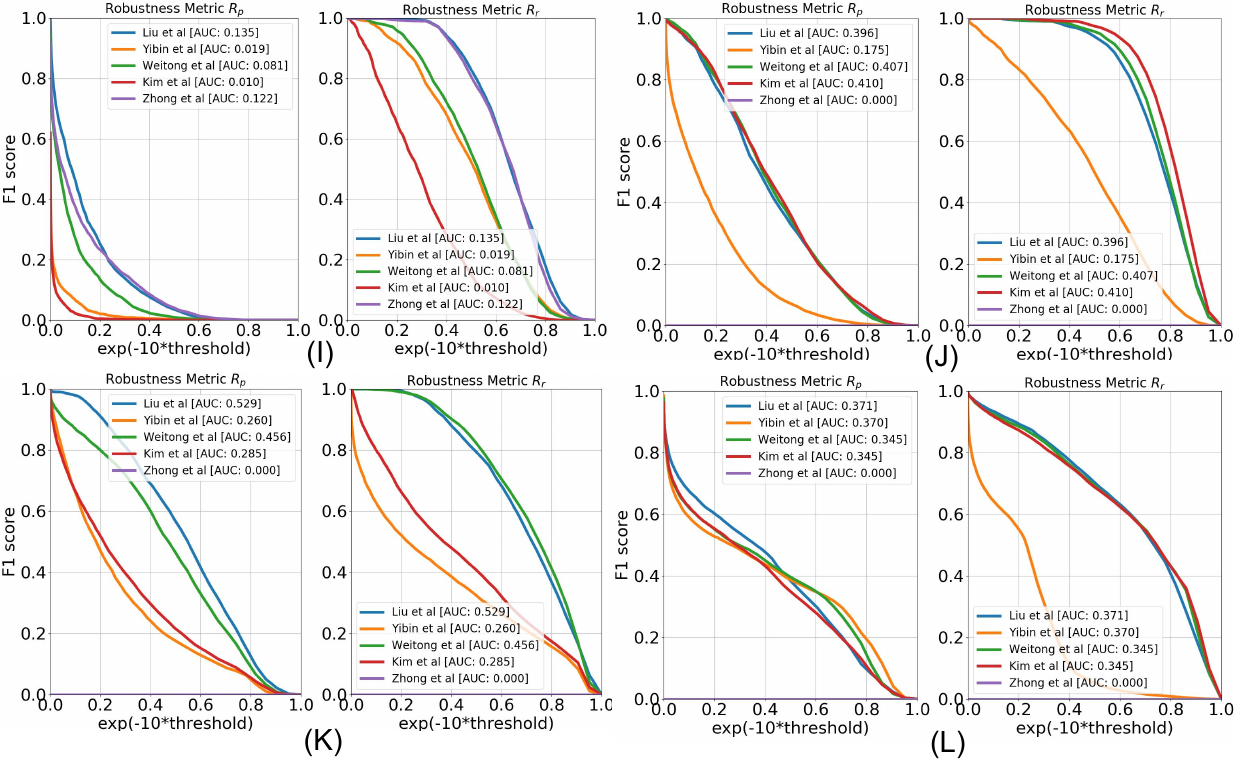}
    \caption{The image, arranged from left to right, displays the robustness metrics \( R_p \) and \( R_r \) from LiDAR track teams: (I) Sewerage, (J) Long Corridor, (K) Multi Floor, (L) Block LiDAR}
    \label{fig:robust_lidar2}
\end{figure*}

\begin{figure*}[!h]
    \centering
    \includegraphics[width=1.0\linewidth,keepaspectratio]{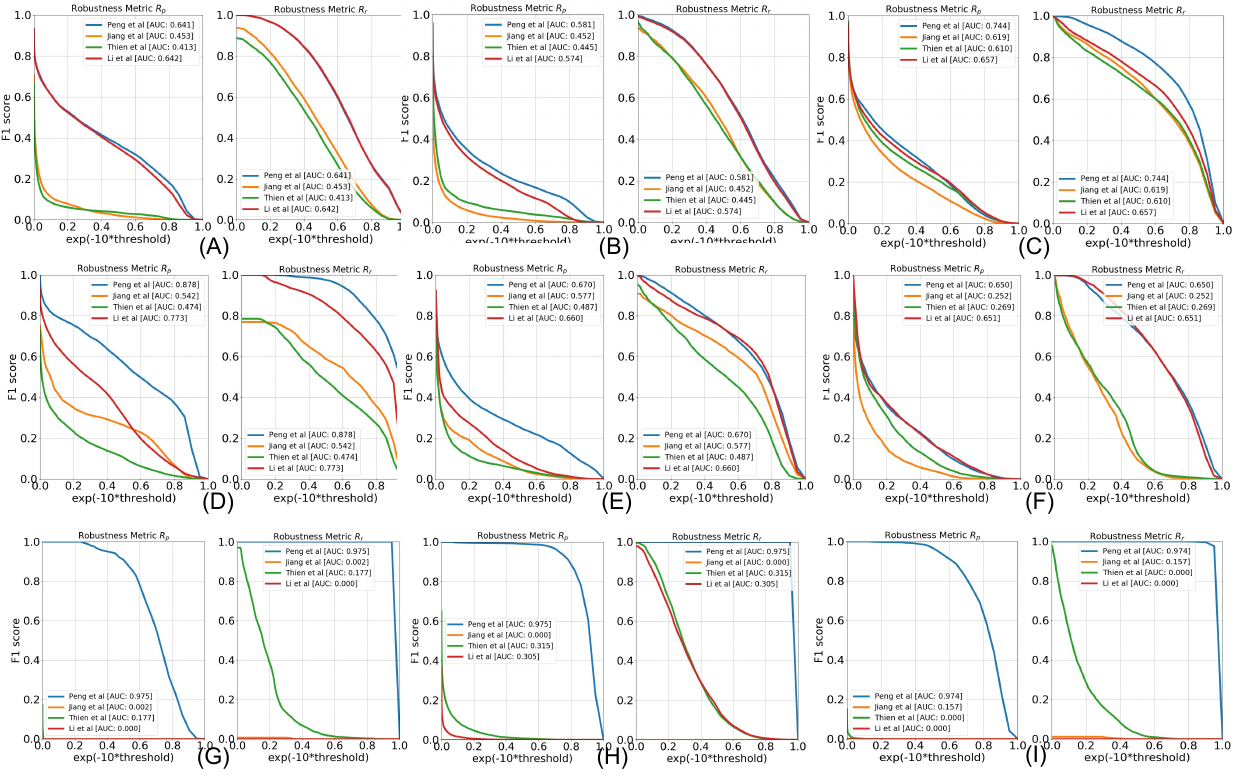}
    \caption{The image, arranged from left to right, displays the robustness metrics \( R_p \) and \( R_r \) from visual track teams: (A) Low Light1, (B) Low Light2, (C) Over Exposure, and (D) Flash Light,(E) Smoke Room (F) Outdoor Night, (G) End of World, (H) Moon (I) Western Desert}
    \label{fig:robust_visual}
\end{figure*}

\subsection{Robustness Evaluation}
In section 4.2, we presented a summary of the robustness metrics for the LiDAR and visual tracks and omitted the detailed robustness plots for each sequence due to the constraints of the paper's length. \fref{fig:robust_lidar1} and \fref{fig:robust_lidar2} illustrate the robustness metrics on each sequence for the LiDAR teams, while \fref{fig:robust_visual} show the results from visual track teams.


\end{document}